\documentclass[journal]{IEEEtran}
\usepackage[latin9]{inputenc}
\usepackage{array}
\usepackage{multirow}
\usepackage{amsmath}
\usepackage{graphicx}
\usepackage{xargs}[2008/03/08]
\PassOptionsToPackage{normalem}{ulem}
\usepackage{ulem}

\makeatletter

\providecommand{\tabularnewline}{\\}

 \let\oldforeign@language\foreign@language
 \DeclareRobustCommand{\foreign@language}[1]{%
   \lowercase{\oldforeign@language{#1}}}

\ifCLASSINFOpdf
\else
\fi

\usepackage{wrapfig}
\usepackage{amssymb}
\usepackage{algorithm, algpseudocode}

\usepackage[english]{babel}
\usepackage{colortbl}

\usepackage{url}
\urldef{\mailsa}\path|{alfred.hofmann, ursula.barth, ingrid.haas, frank.holzwarth,|
\urldef{\mailsb}\path|anna.kramer, leonie.kunz, christine.reiss, nicole.sator,|
\urldef{\mailsc}\path|erika.siebert-cole, peter.strasser, lncs}@springer.com|

\usepackage{graphicx}

\makeatother

\begin{document}
% paper title% Titles are generally capitalized except for words such as a, an, and, as,% at, but, by, for, in, nor, of, on, or, the, to and up, which are usually% not capitalized unless they are the first or last word of the title.% Linebreaks \\ can be used within to get better formatting as desired.% Do not put math or special symbols in the title.

\global\long\def\bW{\boldsymbol{W}}

\global\long\def\be{\boldsymbol{e}}

\global\long\def\modelEBM{\text{EBM}}

\global\long\def\modelRBM{\text{EAD }_{\text{RBM}}}

\global\long\def\modelSRBM{\text{EAD }_{\text{S-RBM}}}

\global\long\def\modelEAD{\text{EAD}}

\global\long\def\modelDBM{\text{EAD }_{\text{DBM}}}

\global\long\def\modelDBMone{\text{EAD }_{\text{DBM-100}}}

\global\long\def\modelDBMtwo{\text{EAD }_{\text{DBM-200}}}

\global\long\def\modelSDBM{\text{EAD }_{\text{S-DBM}}}

\global\long\def\modelSDBMtwo{\text{EAD }_{\text{S-DBM-200}}}

\global\long\def\bA{\boldsymbol{A}}

\global\long\def\bB{\boldsymbol{B}}

\global\long\def\bC{\boldsymbol{C}}

\global\long\def\bc{\boldsymbol{c}}

\global\long\def\bLambda{\boldsymbol{\Lambda}}

\global\long\def\bGamma{\boldsymbol{\Gamma}}

\global\long\def\norm{\left\Vert \right\Vert }

\global\long\def\realmn{\realset^{m\times n}}

\global\long\def\realm{\realset^{m}}

\global\long\def\bb{\boldsymbol{b}}

\global\long\def\rank{\text{rank}}

\global\long\def\lrr#1{\left(#1\right)}

\global\long\def\lrc#1{\left\{  #1\right\}  }

\global\long\def\lra#1{\left\langle #1\right\rangle }

\global\long\def\lrs#1{\left[#1\right]}

\global\long\def\lrv#1{\left|#1\right|}

\global\long\def\mattwtw#1#2#3#4{\left[\begin{array}{rr}
 #1  &  #2\\
 #3  &  #4 
\end{array}\right]}

\global\long\def\mattrtr#1#2#3#4#5#6#7#8#9{\left[\begin{array}{rrr}
 #1  &  #2  &  #3\\
 #4  &  #5  &  #6\\
 #7  &  #8  &  #9 
\end{array}\right]}

\global\long\def\partfunc{\mathcal{Z}}

\global\long\def\model{\text{BNDBM}}

\global\long\def\CDprob{\text{CD }^{prob}}

\global\long\def\r{r}

\global\long\def\rb{\boldsymbol{r}}

\global\long\def\h{h}

\global\long\def\hb{\boldsymbol{h}}

\global\long\def\vb{\boldsymbol{v}}

\global\long\def\v{v}

\global\long\def\xb{\boldsymbol{x}}

\global\long\def\x{x}

\global\long\def\MODEL{\mathrm{MV.RBM}}

\global\long\def\vbias{\mathrm{\boldsymbol{a}}}

\global\long\def\vbiasunit{a}

\global\long\def\visible{\boldsymbol{v}}

\global\long\def\vunit{v}

\global\long\def\vnumunits{M}

\global\long\def\hidden{\mathrm{\boldsymbol{h}}}

\global\long\def\hunit{h}

\global\long\def\hiddenone{\mathrm{\boldsymbol{h}}^{\lrr 1}}

\global\long\def\hiddenoneT{\mathrm{\boldsymbol{h}}^{\lrr 1\top}}

\global\long\def\hiddentwo{\mathrm{\boldsymbol{h}}^{\lrr 2}}

\global\long\def\hiddentwoT{\mathrm{\boldsymbol{h}}^{\lrr 2\top}}

\global\long\def\hbias{\mathrm{\boldsymbol{b}}}

\global\long\def\hbiasunit{\mathrm{b}}

\global\long\def\hbiasone{\mathrm{\boldsymbol{b}}^{\lrr 1}}

\global\long\def\hbiasoneT{\mathrm{\boldsymbol{b}}^{\lrr 1\top}}

\global\long\def\hbiastwo{\mathrm{\boldsymbol{b}}^{\lrr 2}}

\global\long\def\hbiastwoT{\mathrm{\boldsymbol{b}}^{\lrr 2\top}}

\global\long\def\hnumunits{K}

\global\long\def\hnumunitsone{K_{1}}

\global\long\def\hnumunitstwo{K_{2}}

\global\long\def\Wone{\boldsymbol{W}^{\lrr 1}}

\global\long\def\Wtwo{\boldsymbol{W}^{\lrr 2}}

\global\long\def\Wthree{\boldsymbol{W}^{\lrr 3}}

\global\long\def\w{w}

\global\long\def\wone{w^{\lrr 1}}

\global\long\def\wtwo{w^{\lrr 2}}

\global\long\def\paramset{\boldsymbol{\boldsymbol{\Psi}}}

\global\long\def\varparamset{\tilde{\boldsymbol{\Psi}}}

\global\long\def\condmean{\mu}

\global\long\def\varmean{\tilde{\mu}}

\global\long\def\varmeanlayer{\tilde{\boldsymbol{\mu}}}

\global\long\def\sigmoid{\sigma}

\global\long\def\bninput{\boldsymbol{x}}

\global\long\def\bnscale{\gamma}

\global\long\def\bnshift{\beta}

\global\long\def\bnorm{\mathcal{B}}

\global\long\def\bnscalesimple{\text{\ensuremath{\bar{\gamma}}}}

\global\long\def\bnshiftsimple{\text{\ensuremath{\bar{\beta}}}}

\global\long\def\bnparamset{\boldsymbol{\Gamma}}

\global\long\def\bnvarparamset{\tilde{\boldsymbol{\Gamma}}}

\global\long\def\loglike{L}

\global\long\def\bnscalesimplevisible{\bar{\boldsymbol{\lambda}}}

\global\long\def\bnscalevisibleunit{\lambda}

\global\long\def\bnscalesimplevisibleunit{\bar{\lambda}}

\global\long\def\bnshiftsimplevisible{\boldsymbol{\bar{\alpha}}}

\global\long\def\bnshiftvisibleunit{\alpha}

\global\long\def\bnshiftsimplevisibleunit{\bar{\alpha}}

\global\long\def\bnscalesimplehidden{\boldsymbol{\bar{\gamma}}}

\global\long\def\bnscalehiddenunit{\gamma}

\global\long\def\bnscalesimplehiddenunit{\bar{\gamma}}

\global\long\def\bnshiftsimplehidden{\boldsymbol{\bar{\beta}}}

\global\long\def\bnshifthiddenunit{\beta}

\global\long\def\bnshiftsimplehiddenunit{\bar{\beta}}

\global\long\def\bnlinearinputvisible{s}

\global\long\def\bnlinearinputhidden{t}

\global\long\def\elewise{\odot}

\newcommand{\sidenote}[1]{\marginpar{\small \emph{\color{Medium}#1}}}

\global\long\def\se{\hat{\text{se}}}

\global\long\def\interior{\text{int}}

\global\long\def\boundary{\text{bd}}

\global\long\def\ML{\textsf{ML}}

\global\long\def\GML{\mathsf{GML}}

\global\long\def\HMM{\mathsf{HMM}}

\global\long\def\support{\text{supp}}

\global\long\def\new{\text{*}}

\global\long\def\stir{\text{Stirl}}

\global\long\def\mA{\mathcal{A}}

\global\long\def\mB{\mathcal{B}}

\global\long\def\mF{\mathcal{F}}

\global\long\def\mK{\mathcal{K}}

\global\long\def\mH{\mathcal{H}}

\global\long\def\mX{\mathcal{X}}

\global\long\def\mZ{\mathcal{Z}}

\global\long\def\mS{\mathcal{S}}

\global\long\def\Ical{\mathcal{I}}

\global\long\def\mT{\mathcal{T}}

\global\long\def\Pcal{\mathcal{P}}

\global\long\def\dist{d}

\global\long\def\HX{\entro\left(X\right)}
 \global\long\def\entropyX{\HX}

\global\long\def\HY{\entro\left(Y\right)}
 \global\long\def\entropyY{\HY}

\global\long\def\HXY{\entro\left(X,Y\right)}
 \global\long\def\entropyXY{\HXY}

\global\long\def\mutualXY{\mutual\left(X;Y\right)}
 \global\long\def\mutinfoXY{\mutualXY}

\global\long\def\given{\mid}

\global\long\def\gv{\given}

\global\long\def\goto{\rightarrow}

\global\long\def\asgoto{\stackrel{a.s.}{\longrightarrow}}

\global\long\def\pgoto{\stackrel{p}{\longrightarrow}}

\global\long\def\dgoto{\stackrel{d}{\longrightarrow}}

\global\long\def\lik{\mathcal{L}}

\global\long\def\logll{\mathit{l}}

\global\long\def\vectorize#1{\mathbf{#1}}

\global\long\def\vt#1{\mathbf{#1}}

\global\long\def\gvt#1{\boldsymbol{#1}}

\global\long\def\idp{\ \bot\negthickspace\negthickspace\bot\ }
 \global\long\def\cdp{\idp}

\global\long\def\das{:=}

\global\long\def\id{\mathbb{I}}

\global\long\def\idarg#1#2{\id\left\{  #1,#2\right\}  }

\global\long\def\iid{\stackrel{\text{iid}}{\sim}}

\global\long\def\bzero{\vt 0}

\global\long\def\bone{\mathbf{1}}

\global\long\def\boldm{\boldsymbol{m}}

\global\long\def\bff{\vt f}

\global\long\def\bx{\boldsymbol{x}}

\global\long\def\bl{\boldsymbol{l}}

\global\long\def\bu{\boldsymbol{u}}

\global\long\def\ba{\boldsymbol{a}}

\global\long\def\bo{\boldsymbol{o}}

\global\long\def\bh{\boldsymbol{h}}

\global\long\def\bs{\boldsymbol{s}}

\global\long\def\bz{\boldsymbol{z}}

\global\long\def\xnew{y}

\global\long\def\bxnew{\boldsymbol{y}}

\global\long\def\bX{\boldsymbol{X}}

\global\long\def\tbx{\tilde{\bx}}

\global\long\def\by{\boldsymbol{y}}

\global\long\def\bY{\boldsymbol{Y}}

\global\long\def\bZ{\boldsymbol{Z}}

\global\long\def\bU{\boldsymbol{U}}

\global\long\def\bv{\boldsymbol{v}}

\global\long\def\bn{\boldsymbol{n}}

\global\long\def\bV{\boldsymbol{V}}

\global\long\def\bI{\boldsymbol{I}}

\global\long\def\bw{\vt w}

\global\long\def\balpha{\gvt{\alpha}}

\global\long\def\bbeta{\gvt{\beta}}

\global\long\def\bmu{\gvt{\mu}}

\global\long\def\btheta{\boldsymbol{\theta}}

\global\long\def\bdelta{\boldsymbol{\delta}}

\global\long\def\blambda{\boldsymbol{\lambda}}

\global\long\def\bgamma{\boldsymbol{\gamma}}

\global\long\def\bpsi{\boldsymbol{\psi}}

\global\long\def\bphi{\boldsymbol{\phi}}

\global\long\def\bpi{\boldsymbol{\pi}}

\global\long\def\bomega{\boldsymbol{\omega}}

\global\long\def\bepsilon{\boldsymbol{\epsilon}}

\global\long\def\btau{\boldsymbol{\tau}}

\global\long\def\realset{\mathbb{R}}

\global\long\def\realn{\realset^{n}}

\global\long\def\integerset{\mathbb{Z}}

\global\long\def\natset{\integerset}

\global\long\def\integer{\integerset}

\global\long\def\natn{\natset^{n}}

\global\long\def\rational{\mathbb{Q}}

\global\long\def\rationaln{\rational^{n}}

\global\long\def\complexset{\mathbb{C}}

\global\long\def\comp{\complexset}

\global\long\def\compl#1{#1^{\text{c}}}

\global\long\def\and{\cap}

\global\long\def\compn{\comp^{n}}

\global\long\def\comb#1#2{\left({#1\atop #2}\right) }

\global\long\def\nchoosek#1#2{\left({#1\atop #2}\right)}

\global\long\def\param{\vt w}

\global\long\def\Param{\Theta}

\global\long\def\meanparam{\gvt{\mu}}

\global\long\def\Meanparam{\mathcal{M}}

\global\long\def\meanmap{\mathbf{m}}

\global\long\def\logpart{A}

\global\long\def\simplex{\Delta}

\global\long\def\simplexn{\simplex^{n}}

\global\long\def\dirproc{\text{DP}}

\global\long\def\ggproc{\text{GG}}

\global\long\def\DP{\text{DP}}

\global\long\def\ndp{\text{nDP}}

\global\long\def\hdp{\text{HDP}}

\global\long\def\gempdf{\text{GEM}}

\global\long\def\rfs{\text{RFS}}

\global\long\def\bernrfs{\text{BernoulliRFS}}

\global\long\def\poissrfs{\text{PoissonRFS}}

\global\long\def\grad{\gradient}
 \global\long\def\gradient{\nabla}

\global\long\def\partdev#1#2{\partialdev{#1}{#2}}
 \global\long\def\partialdev#1#2{\frac{\partial#1}{\partial#2}}

\global\long\def\partddev#1#2{\partialdevdev{#1}{#2}}
 \global\long\def\partialdevdev#1#2{\frac{\partial^{2}#1}{\partial#2\partial#2^{\top}}}

\global\long\def\closure{\text{cl}}

\global\long\def\cpr#1#2{\Pr\left(#1\ |\ #2\right)}

\global\long\def\var{\text{Var}}

\global\long\def\Var#1{\text{Var}\left[#1\right]}

\global\long\def\cov{\text{Cov}}

\global\long\def\Cov#1{\cov\left[ #1 \right]}

\global\long\def\COV#1#2{\underset{#2}{\cov}\left[ #1 \right]}

\global\long\def\corr{\text{Corr}}

\global\long\def\sst{\text{T}}

\global\long\def\SST{\sst}

\global\long\def\ess{\mathbb{E}}

\global\long\def\Ess#1{\ess\left[#1\right]}

\newcommandx\ESS[2][usedefault, addprefix=\global, 1=]{\underset{#2}{\ess}\left[#1\right]}

\global\long\def\fisher{\mathcal{F}}

\global\long\def\bfield{\mathcal{B}}
 \global\long\def\borel{\mathcal{B}}

\global\long\def\bernpdf{\text{Bernoulli}}

\global\long\def\betapdf{\text{Beta}}

\global\long\def\dirpdf{\text{Dir}}

\global\long\def\gammapdf{\text{Gamma}}

\global\long\def\gaussden#1#2{\text{Normal}\left(#1, #2 \right) }

\global\long\def\gauss{\mathbf{N}}

\global\long\def\gausspdf#1#2#3{\text{Normal}\left( #1 \lcabra{#2, #3}\right) }

\global\long\def\multpdf{\text{Mult}}

\global\long\def\poiss{\text{Pois}}

\global\long\def\poissonpdf{\text{Poisson}}

\global\long\def\pgpdf{\text{PG}}

\global\long\def\wshpdf{\text{Wish}}

\global\long\def\iwshpdf{\text{InvWish}}

\global\long\def\nwpdf{\text{NW}}

\global\long\def\niwpdf{\text{NIW}}

\global\long\def\studentpdf{\text{Student}}

\global\long\def\unipdf{\text{Uni}}

\global\long\def\transp#1{\transpose{#1}}
 \global\long\def\transpose#1{#1^{\mathsf{T}}}

\global\long\def\mgt{\succ}

\global\long\def\mge{\succeq}

\global\long\def\idenmat{\mathbf{I}}

\global\long\def\trace{\mathrm{tr}}

\global\long\def\argmax#1{\underset{_{#1}}{\text{argmax}} }

\global\long\def\argmin#1{\underset{_{#1}}{\text{argmin}\ } }

\global\long\def\diag{\text{diag}}

\global\long\def\norm{}

\global\long\def\spn{\text{span}}

\global\long\def\vtspace{\mathcal{V}}

\global\long\def\field{\mathcal{F}}
 \global\long\def\ffield{\mathcal{F}}

\global\long\def\inner#1#2{\left\langle #1,#2\right\rangle }
 \global\long\def\iprod#1#2{\inner{#1}{#2}}

\global\long\def\dprod#1#2{#1 \cdot#2}

\global\long\def\norm#1{\left\Vert #1\right\Vert }

\global\long\def\entro{\mathbb{H}}

\global\long\def\entropy{\mathbb{H}}

\global\long\def\Entro#1{\entro\left[#1\right]}

\global\long\def\Entropy#1{\Entro{#1}}

\global\long\def\mutinfo{\mathbb{I}}

\global\long\def\relH{\mathit{D}}

\global\long\def\reldiv#1#2{\relH\left(#1||#2\right)}

\global\long\def\KL{KL}

\global\long\def\KLdiv#1#2{\KL\left(#1\parallel#2\right)}
 \global\long\def\KLdivergence#1#2{\KL\left(#1\ \parallel\ #2\right)}

\global\long\def\crossH{\mathcal{C}}
 \global\long\def\crossentropy{\mathcal{C}}

\global\long\def\crossHxy#1#2{\crossentropy\left(#1\parallel#2\right)}

\global\long\def\breg{\text{BD}}

\global\long\def\lcabra#1{\left|#1\right.}

\global\long\def\lbra#1{\lcabra{#1}}

\global\long\def\rcabra#1{\left.#1\right|}

\global\long\def\rbra#1{\rcabra{#1}}

\global\long\def\likelihood{\mathcal{L}}

\global\long\def\normal{\mathcal{N}}

\global\long\def\bSigma{\boldsymbol{\Sigma}}

\global\long\def\dataset{\mathcal{D}}

\title{Detection of Unknown Anomalies in Streaming Videos with Generative
Energy-based Boltzmann Models}

\author{Hung~Vu,~Tu~Dinh~Nguyen~and~Dinh~Phung\thanks{Hung Vu, Tu Dinh Nguyen and Dinh Phung are with the Center for Pattern
Recognition and Data Analytics, School of Information Technology,
Deakin University, Geelong, Australia. e-mails: hungv@deakin.edu.au,
tu.nguyen@deakin.edu.au, dinh.phung@deakin.edu.au.\protect \\
\par\emph{This work is under consideration at Pattern Recognition
Letters.}}}

\maketitle
$\maketitle$

\author{}

% author names and IEEE memberships% note positions of commas and nonbreaking spaces ( ~ ) LaTeX will not break% a structure at a ~ so this keeps an author's name from being broken across% two lines.% use \thanks{} to gain access to the first footnote area% a separate \thanks must be used for each paragraph as LaTeX2e's \thanks% was not built to handle multiple paragraphs

\author{% <-this % stops a space\thanks{M. Shell was with the Department of Electrical and Computer Engineering,
Georgia Institute of Technology, Atlanta, GA, 30332 USA e-mail: (see
http://www.michaelshell.org/contact.html).}% <-this % stops a space\thanks{J. Doe and J. Doe are with Anonymous University.}% <-this % stops a space\thanks{Manuscript received April 19, 2005; revised August 26, 2015.}}

% note the % following the last \IEEEmembership and also \thanks - % these prevent an unwanted space from occurring between the last author name% and the end of the author line. i.e., if you had this:% % \author{....lastname \thanks{...} \thanks{...} }%                     ^------------^------------^----Do not want these spaces!% a space would be appended to the last name and could cause every name on that% line to be shifted left slightly. This is one of those "LaTeX things". For% instance, "\textbf{A} \textbf{B}" will typeset as "A B" not "AB". To get% "AB" then you have to do: "\textbf{A}\textbf{B}"% \thanks is no different in this regard, so shield the last } of each \thanks% that ends a line with a % and do not let a space in before the next \thanks.% Spaces after \IEEEmembership other than the last one are OK (and needed) as% you are supposed to have spaces between the names. For what it is worth,% this is a minor point as most people would not even notice if the said evil% space somehow managed to creep in.


\markboth{The paper is under consideration at Pattern Recognition Letters}{Shell \MakeLowercase{\textit{et al.}}: Bare Demo of IEEEtran.cls
for IEEE Journals}

\begin{abstract}
Abnormal event detection is one of the important objectives in research
and practical applications of video surveillance. However, there are
still three challenging problems for most anomaly detection systems
in practical setting: limited labeled data, ambiguous definition of
``abnormal'' and expensive feature engineering steps. This paper
introduces a unified detection framework to handle these challenges
using energy-based models, which are powerful tools for unsupervised
representation learning. Our proposed models are firstly trained on
unlabeled raw pixels of image frames from an input video rather than
hand-crafted visual features; and then identify the locations of abnormal
objects based on the errors between the input video and its reconstruction
produced by the models. To handle video stream, we develop an online
version of our framework, wherein the model parameters are updated
incrementally with the image frames arriving on the fly. Our experiments
show that our detectors, using Restricted Boltzmann Machines (RBMs)
and Deep Boltzmann Machines (DBMs) as core modules, achieve superior
anomaly detection performance to unsupervised baselines and obtain
accuracy comparable with the state-of-the-art approaches when evaluating
at the pixel-level. More importantly, we discover that our system
trained with DBMs is able to simultaneously perform scene clustering
and scene reconstruction. This capacity not only distinguishes our
method from other existing detectors but also offers a unique tool
to investigate and understand how the model works. 
\end{abstract}


% For peer review papers, you can put extra information on the cover% page as needed:% \ifCLASSOPTIONpeerreview% \begin{center} \bfseries EDICS Category: 3-BBND \end{center}% \fi% For peerreview papers, this IEEEtran command inserts a page break and% creates the second title. It will be ignored for other modes.

\IEEEpeerreviewmaketitle{}

\section{Introduction\label{sec:Introduction}}

In the last few years, the security and safety concerns in public
places and restricted areas have increased the need for visual surveillance.
Large distributed networks of many high quality cameras have been
deployed and producing an enormous amount of data every second. Monitoring
and processing such huge information manually are infeasible in practical
applications. As a result, it is imperative to develop autonomous
systems that can identify, highlight, predict anomalous objects or
events, and then help to make early interventions to prevent hazardous
actions (e.g., fighting or a stranger dropping a suspicious case)
or unexpected accidents (e.g., falling or a wrong movement on one-way
streets). Video anomaly detection can also be widely-used in variety
of applications such as restricted-area surveillance, traffic analysis,
group activity detection, home security to name a few. The recent
studies \cite{AngelaSodemann-2012-Review} show that video anomaly
detection has received considerable attention in the research community
and become one of the essential problems in computer vision. However,
deploying surveillance systems in real-world applications poses three
main challenges: a) the easy availability of unlabeled data but lack
of labeled training data; b) no explicit definition of anomaly in
real-life video surveillance and c) expensive hand-crafted feature
extraction exacerbated by the increasing complexity in videos. 

The first challenge comes from the fast growing availability of low-cost
surveillance cameras nowadays. A typical RGB camera with the resolution
of $340\times640$ pixels can add more than one terabyte video data
every day. To label this data, an annotation process is required to
produce a ground-truth mask for every video frame. In particular,
a person views the video, stops at a frame and then assigns pixel
regions as anomalous objects or behaviors wherever applicable. This
person has to be well-trained and carefully look at every single detail
all the time, otherwise he might miss some unusual events that suddenly
appear. This process is extremely labor-intensive, rendering it impossible
to obtain large amount of labeled data; and hence upraising the demand
for a method that can exploit the overabundant unlabeled videos rather
than relying on the annotated one.

The second challenge of no explicit definition is due to the diversity
of abnormal events in reality. In some constrained environments, abnormalities
are well-defined, for example, putting goods into pocket in the supermarket
\cite{Zhou.G_Wu.Y_2009InterConfonInfoEngCS_supervised}; hence we
can view the problem as activity recognition and apply a machine learning
classifier to detect suspicious behaviors. However, anomaly objects
in most scenarios are undefined, e.g., any objects except for cars
on free-way can be treated as anomaly. Therefore, an anomaly detection
algorithm faces the fact that it has scarce information about what
it needs to predict until they actually appear. As a result, developing
a good anomaly detector to detect \emph{unknown} anomalous objects
is a very challenging problem. 

Last but not least, most anomaly detectors normally rely on hand-crafted
features such as Histogram of Oriented Gradients (HOG) \cite{Dalal.Navneet_Triggs.Bill_CVPR2005_HOG},
Histogram of Optical Flow (HOF) \cite{Dalal.Navneet_etal_ECCV2006_HOF}
or Optical Flow \cite{Lucas.Bruce_Kanade.Takeo_IJCAI1981_OpticalFlow}
to perform well. These features were carefully designed using a number
of trail-and-error experiments from computer vision community over
many years. However, these good features are known to have expensive
computation and expert knowledge dependence. Moreover, a feature extraction
procedure should be redesigned or modified to adapt to the purpose
of each particular application.

To that end, we introduce a novel energy-based framework to tackle
all aforementioned challenges in anomaly detection. Our proposed system,
termed Energy-based Anomaly Detector ($\modelEAD$), is trained in
completely unsupervised learning manner to model the complex distribution
of data, and thus captures the data regularity and variations. The
learning requires neither the label information nor an explicit definition
of abnormality that is assumed to be the irregularity in the data
\cite{AngelaSodemann-2012-Review}, hence effectively addressing the
first two challenges. In addition, our model works directly on the
raw pixels at the input layer, and transforms the data to hierarchical
representations at higher layers using an efficient inference scheme
\cite{Salakhutdinov.Ruslan_Hinton.Geoffrey_2009AISTATS,Nguyen.TuDinh_eltal_2013ACML_NRBM,Tran.Truyen_etal_2015JournalBI_eNRBM,Nguyen.TuDinh_etal_2013PAKDD_Mv.RBM_latenpatient}.
These representations are more compact, reflects the underlying factors
in the data well, and can be effectively used for further tasks. Therefore
our proposed framework can bypass the third challenge of expensive
feature engineering requirement.

In order to build our system, we first rescale the video into different
resolutions to handle objects of varying sizes. At each resolution,
the video frames are partitioned into overlapping patches, which are
then gathered into groups of the same location in the frame. The energy-based
module is then trained on these groups, and used to reconstruct the
input data at the detection stage once the training has finished.
An image patch is identified as a \emph{potential} candidate residing
in an abnormal region if its reconstruction error is larger than a
predefined threshold. Next we find the connected components of these
candidates spanning over a fixed number of frames to finally obtain
abnormal objects.

To build the energy-based module for our system, our previous attempt
\cite{hungvu_etal_2017PAKDD_anomaly_RBM} used Restricted Boltzmann
Machines (RBMs) \cite{Freund_Haussler_1994_unsupervised,Hinton.Geoffrey_2002NeuralComp_CD},
an expressive class of two-layer generative networks; we named this
version $\modelRBM$. Our $\modelRBM$ first employs a single RBM
to cluster similar image patches into groups, and then builds an independent
RBM for each group. This framework shows promising detection results;
however, one limitation is that it is a complicated multi-stage system
which requires to maintain two separate modules with a number of RBM
models for clustering and reconstruction tasks.

To address this problem, we seek for a simpler system that can perform
both tasks using only a single model. We investigate the hierarchical
structure in the video data, and observe that the fine-detailed representations
are rendered at low levels whilst the group property is described
at higher, more abstract levels. Based on these observations, we further
introduce the second version of our framework that employs Deep Boltzmann
Machines (DBMs) \cite{Salakhutdinov.Ruslan_Hinton.Geoffrey_2009AISTATS}
as core modules, termed $\modelDBM$. Instead of using many shallow
RBM models, this version uses only one deep multi-layer DBM architecture,
wherein each layer has responsibility for clustering or reconstructing
the data. Whilst keeping the capacity of unsupervised learning, automated
representation learning, detecting unknown localized abnormalities
for both offline and streaming settings as in $\modelRBM$, the $\modelDBM$
offers two more advanced features. Firstly, it is a \emph{unified}
framework that can handle all the stages of modeling, clustering and
localizing to detect from the beginning to the end. The second feature
is the data and model interpretability at abstract levels. Most existing
systems can detect anomaly with high performance, but they fail to
provide any explanation of why such detection is obtained. By contrast,
we demonstrate that our $\modelDBM$ is able to understand the scene,
show the reason why it makes fault alarms, and hence our detection
results are completely explainable. This property is especially useful
for debugging during the system development and error diagnostics
during the deployment. To the best of our knowledge, our work is the
first one that uses DBM for anomaly detection in video data, and also
the first work in DBM's literature using a single model for both clustering
and reconstructing data. Thus, we believe that our system stands out
among most existing methods and offers an alternative approach in
anomaly detection research.

We conduct comprehensive experiments on three benchmark datasets:
UCSD Ped 1, Ped 2 and Avenue using a number of evaluation metrics.
The results show that our single-model $\modelDBM$ obtains equivalent
performances to multi-model $\modelRBM$, whilst it can detect abnormal
objects more accurately than standard baselines and achieve competitive
results with those of state-of-the-art approaches.

The rest of the paper is organized as follows. Sec.~\ref{sec:Related-work}
discusses the related work whilst Sec.~\ref{sec:Energy-based-Model}
presents an introduction to RBM and DBM. Two variants of our anomaly
detection systems, $\modelRBM$ and $\modelDBM$, are described in
Sec.~\ref{sec:Framework} followed by experiments reported in Sec.~\ref{sec:Experiment}.
Finally, Sec.~\ref{sec:Conclusion} concludes the paper.

\section{Related work\label{sec:Related-work}}

To date, many attempts have been proposed to build up video anomaly
detection systems \cite{AngelaSodemann-2012-Review}. Two typical
approaches are: supervised methods that use the labels to cast anomaly
detection problem to binary or one-class classification problems;
and unsupervised methods that learn to generalize the data without
labels, and hence can discover irregularity afterwards. In this section,
we provide a brief overview of models in these two approaches before
discussing the recent lines of deep learning and energy-based work
for video anomaly detection.

The common solution in the supervised approach is to train binary
classifiers on both abnormal and normal data. \cite{Cui.Xinyi_etal_2011CVPR_supervised_energybased}
firstly extracts combined features of interaction energy potentials
and optical flows at every interest point before training Support
Vector Machines (SVM) on bag-of-word representation of such features.
\cite{Singh.Dinesh_Mohan.Krishna_2017PR_Graphbased} use a binary
classifier on the bag-of-graph constructed from Space-Time Interest
Points (STIP) descriptors \cite{Laptev.Ivan_IJCV2005_STIP}. Another
approach is to ignore the abnormal data, and use normal data only
to train the models. For example, Support Vector Data Description
(SVDD) \cite{Zhang.Ying_etal_2016PR_Supervised_MotionAppearance}
first learns the spherical boundary for normal data, and then identifies
unusual events based on the distances from such events to the boundary.
Sparse Coding \cite{Lu.Cewu_etal_2013ICCV_Supervised-Sparse} and
Locality-Constrained Affine Subspace Coding \cite{Fan.Yaxiang_etal_2017JEI_Supervised_LocalityConstrainedAffineSubspaceCoding}
assume that regular examples can be presented via a learned dictionary
whilst irregular events usually cause high reconstruction errors,
and thus can be separated from the regular ones. Several methods such
as Chaotic Invariant \cite{Wu.Shandon_etal_2010CVPR_supervised_chaotic}
are based on mixture models to learn the probability distribution
of regular data and estimate the probability of an observation to
be abnormal for anomaly detection. Overall, all methods in the supervised
approach require labor-intensive annotation process, rendering them
less applicable in practical large-scale applications.

The unsupervised approach offers an appealing way to train models
without the need for labeled data. The typical strategy is to capture
the majority of training data points that are assumed to be normal
examples. One can first split a video frame into a grid and use optical
flow counts over grid cells as feature vectors \cite{Pham.DucSon_etal_2011ICDM_principle_eigenvectors}.
Next the Principal Component Analysis works on these vectors to find
a lower dimensional principal subspace that containing the most information
of the data, and then projecting the data onto the complement residual
subspace to compute the residual signals. Higher signals indicate
more suspicious data points. Sparse Coding, besides being used in
supervised learning as above, is also applied in unsupervised manner
wherein feature vectors are HOG or HOF descriptors of points of interest
inside spatio-temporal volumes \cite{Zhao.Bin_etal_2011CVPR_unsupervised-SparseCoding}.
Another way to capture the domination of normality is to train One-Class
SVM (OC-SVM) on the covariance matrix of optical flows and partial
derivatives of connective frames or image patches \cite{Wang.Tian_etal_2017MTA_OpticalFlow_CovMatrix}.
Clustering-based method \cite{Roshtkhari.MehrsanJavan_Levine.MartinD_2013CVPR_unsupervised_streaming_BOV}
encodes regular examples as codewords in bag-of-video-word models.
An ensemble of spatio-temporal volumes is then specified as abnormality
if it is considerably different from the learned codewords. To detect
abnormality for a period in human activity videos, \cite{Duong.ThiV_etal_2005CVPR_HSMM}
introduces Switching Hidden Semi-Markov Model (S-HSMM) based on comparing
the probabilities of normality and abnormality in such period.

All aforementioned unsupervised methods, however, usually rely on
hand-crafted features, such as gradients \cite{Roshtkhari.MehrsanJavan_Levine.MartinD_2013CVPR_unsupervised_streaming_BOV},
HOG \cite{Zhao.Bin_etal_2011CVPR_unsupervised-SparseCoding}, HOF
\cite{Zhao.Bin_etal_2011CVPR_unsupervised-SparseCoding}, optical
flow based features \cite{Pham.DucSon_etal_2011ICDM_principle_eigenvectors,Wang.Tian_etal_2017MTA_OpticalFlow_CovMatrix}.
In recent years, the tremendous success of deep learning in various
areas of computer vision \cite{Guo.Yanming_etal_2016NeuralComp_DL_CV_review}
has motivated a series of studies exploring deep learning techniques
in video anomaly detection. Many deep networks have been used to build
up both supervised anomaly detection frameworks such as Convolutional
Neural Networks (CNN) \cite{Sabokrou.Mohammad_etal_2016CVIU_CNN},
Generative Adversarial Nets (GAN) \cite{Ravanbakhsh.Mahdyar_etal_2017ICIP_DL_GAN},
Convolutional Winner-Take-All Autoencoders \cite{Tran.HTM_Hogg.DC_2017BMVC_DL_ConvWinnerTakeAllAE}
and unsupervised systems such as Convolutional Long-Short Term Memories
\cite{Chong.YongShean_Tay.YongHaur_2017AdvNN_DL_ConvAE_ConvLSTM,Luo.W_etal_2017ICME_DL_ConvLSTMAE,Medel.JeffersonRyan_Savakis.AndreasE_2016CoRR_DL_LSTM},
Convolutional Autoencoders \cite{Chong.YongShean_Tay.YongHaur_2017AdvNN_DL_ConvAE_ConvLSTM,Luo.W_etal_2017ICME_DL_ConvLSTMAE,Ribeiro.Manasses_etal_2017PRL_DL_ConvAE,Hasan.Mahmudul_etla_2016CVPR_DL_ConvAE},
Stacked Denoising Autoencoders \cite{Xu.Dan_etal_2017CVIU_DL_StackedDenoisingAE_OCSVM}.
By focusing on unsupervised learning methods, in what follows we will
give a brief review of the unsupervised deep networks.

By viewing anomaly detection as a reconstruction problem, Hasan et
al. \cite{Hasan.Mahmudul_etla_2016CVPR_DL_ConvAE} proposed to learn
a Convolutional Autoencoder to reconstruct input videos. They show
that a deep architecture with 12 layers trained on raw pixel data
can produce meaningful features comparable with the state-of-the-art
hand-crafted features of HOG, HOF and improved trajectories for video
anomaly detection. \cite{Ribeiro.Manasses_etal_2017PRL_DL_ConvAE}
extends this work by integrating multiple channels of information,
i.e., raw pixels, edges and optical flows, into the network to obtain
better performance. Appearance and Motion Deep Nets (AMDNs) \cite{Xu.Dan_etal_2017CVIU_DL_StackedDenoisingAE_OCSVM}
is a fusion framework to encode both appearance and motion information
in videos. Three Stacked Denoising Autoencoders are constructed on
each type of information (raw patches and optical flows) and their
combination. Each OC-SVM is individually trained on the encoded values
of each network and their decisions are lately fused to form a final
abnormality map. To detect anomaly events across the dimension of
time, \cite{Medel.JeffersonRyan_Savakis.AndreasE_2016CoRR_DL_LSTM}
introduces a Composite Convolutional Long-Short Term Memories (Composite
ConvLSTM) that consists of one encoder and two decoders of past reconstruction
and future prediction. The performance of this network is shown to
be comparable with ConvAE \cite{Hasan.Mahmudul_etla_2016CVPR_DL_ConvAE}.
Several studies \cite{Chong.YongShean_Tay.YongHaur_2017AdvNN_DL_ConvAE_ConvLSTM,Luo.W_etal_2017ICME_DL_ConvLSTMAE}
attempt to combine both ConvAE and ConvLSTM into the same system where
ConvAE has responsibility to capture spatial information whilst temporal
information is learned by ConvLSTM. 

Although deep learning is famous for its capacity of feature learning,
not all aforementioned deep systems utilize this powerful capacity,
for example, the systems in \cite{Ribeiro.Manasses_etal_2017PRL_DL_ConvAE,Xu.Dan_etal_2017CVIU_DL_StackedDenoisingAE_OCSVM}
still depend on hand-crafted features in their designs.  Since we
are interested in deep systems with the capacity of feature learning,
we consider unsupervised deep detectors working directly on raw data
as our closely related work, for example, Hasan et al.'s system \cite{Hasan.Mahmudul_etla_2016CVPR_DL_ConvAE},
CAE \cite{Ribeiro.Manasses_etal_2017PRL_DL_ConvAE}, Composite ConvLSTM
\cite{Medel.JeffersonRyan_Savakis.AndreasE_2016CoRR_DL_LSTM}, ConvLSTM-AE
\cite{Chong.YongShean_Tay.YongHaur_2017AdvNN_DL_ConvAE_ConvLSTM}
and Lu et al's system \cite{Luo.W_etal_2017ICME_DL_ConvLSTMAE}. However,
these detectors are basically trained with the principle of minimizing
\emph{reconstruction loss functions} instead of learning real data
distributions. Low reconstruction error in these systems does not
mean a good model quality because of overfitting problem. As a result,
these methods do not have enough capacity of generalization and do
not reflect the diversity of normality in reality. 

Our proposed methods are based on energy-based models, versatile
frameworks that have rigorous theory in modeling data distributions.
In what follows, we give an overview of energy-based networks that
have been used to solve anomaly detection in general and video anomaly
detection in particular. Restricted Boltzmann Machines (RBMs) are
one of the fundamental energy-based networks with one visible layer
and one hidden layer. In \cite{Do.Kien_etal_2016ADMA_RBM_anomalydetection},
its variant for mixed data is used to detect outliers that are significantly
different from the majority. The free-energy function of RBMs is considered
as an outlier scoring method to separate the outliers from the data.
Another energy-based network to detect anomaly objects is Deep Structured
Energy-based Models (DSEBMs) \cite{Balcan.MariaFlorina_Weinberger.KilianQ_2016PMLR_DL_Energybased}.
DSEBMs are a variant of RBMs with a redefined energy function as the
output of a deterministic deep neural network. Since DSEBMs are trained
with Score Matching \cite{Hyvarinen.Aapo_2005JMLR_scorematching},
they are essentially equivalent to one layer Denoising Autoencoders
\cite{Vincent.Pascal_2011NeuralComputing_ConnectionScoreMatchingAndDAEs}.
 For video anomaly detection, Revathi and Kumar \cite{Revathi.AR_Kumar.Dhananjay_2016_supervised_DBN}
proposed a supervised system of four modules: background estimation,
object segmentation, feature extraction and activity recognition.
The last module of classifying a tracked object to be abnormal or
normal is a deep network trained with DBNs and fine-tuned using a
back-propagation algorithm. Overall, these energy-based detectors
mainly focus on shallow networks, such as RBMs, or the stack of these
networks, i.e., DBNs, but have not investigated the power of deep
energy-based networks, for example, Deep Boltzmann Machines.  For
this reason, we believe that our energy-based video anomaly detectors
are distinct and stand out from other existing frameworks in the literature.

\textbf{}\textbf{}\textbf{}

\section{Energy-based Model\label{sec:Energy-based-Model}}

Energy-based models (EBMs) are a rich family of probabilistic models
that capture the dependencies among random variables. Let us consider
a model with two sets of visible variables $\visible$ and hidden
variables $\bh$ and a parameter set $\paramset$. The idea is to
associate each configuration of all variables with an energy value.
More specifically, the EBM assigns an energy function $E\lrr{\visible,\bh;\paramset}$
for a joint configuration of $\visible$ and $\bh$ and then admits
a Boltzmann distribution (also known as Gibbs distribution) as follows:\vspace{-0.6cm}

\begin{align}
p\lrr{\visible,\bh;\paramset} & =\frac{e^{-E\lrr{\visible,\bh;\paramset}}}{\partfunc\lrr{\paramset}}\label{eq:ebm_joint_distribution}
\end{align}
wherein $\partfunc\lrr{\paramset}=\sum_{\visible,\bh}e^{-E\lrr{\visible,\bh;\paramset}}$
is the normalization constant, also called the partition function.
This guarantees that the $p\lrr{\visible,\bh;\paramset}$ is a proper
density function (p.d.f) wherein the p.d.f is positive and its sum
over space equals to 1.

The learning of energy-based model aims to seek for an optimal parameter
set that assigns the lowest energies (the highest probabilities) to
the training set of $N$ samples: $\mathcal{D}=\lrc{\visible^{\lrs n}}_{n=1}^{N}$.
To that end, the EBM attempts to maximize the data log-likelihood
$\log\mathcal{\loglike}\lrr{\visible;\paramset}=\log\sum_{\bh}p\lrr{\visible,\bh;\paramset}$.
Since the distribution in Eq.~(\ref{eq:ebm_joint_distribution})
can viewed as a member of exponential family, the gradient of log-likelihood
function with respect to parameter $\paramset$ can be derived as:
\begin{align}
\grad_{\paramset}\log\mathcal{L} & =\ess_{\text{data}}\lrs{-\frac{\partial E}{\partial\paramset}}-\ess_{\text{model}}\lrs{-\frac{\partial E}{\partial\paramset}}\label{eq:ebm_gradient_log_like}
\end{align}
Thus the parameters can be updated using the following rule:
\begin{align}
\paramset & =\paramset+\eta\lrr{\ess_{\text{data}}\lrs{-\frac{\partial E}{\partial\paramset}}-\ess_{\text{model}}\lrs{-\frac{\partial E}{\partial\paramset}}}\label{eq:ebm_update}
\end{align}
for a learning rate $\eta>0$. Here $\ess_{\text{data}}$ and $\ess_{\text{model}}$
represent the expectations of partial derivatives over data distribution
and model distribution respectively. Computing these two statistics
are generally intractable, hence we must resort to approximate approaches
such as variational inference \cite{Salakhutdinov.Ruslan_Hinton.Geoffrey_2012_DBM}
or sampling \cite{Hinton.Geoffrey_2002NeuralComp_CD,Tieleman.Tijmen_2008ICML_PCD}. 

In what follows we describe two typical examples of EBMs: Restricted
Boltzmann Machines and Deep Boltzmann Machines that are the core modules
of our proposed anomaly detection systems.

\subsection{Restricted Boltzmann Machines}

Restricted Boltzmann Machine (RBM) \cite{Freund_Haussler_1994_unsupervised,Hinton.Geoffrey_2002NeuralComp_CD}
is a bipartite undirected network with $M$ \emph{binary} visible
units $\visible=\left[\vunit_{1},\vunit_{2},\text{\dots},\vunit_{M}\right]^{\top}\in\{0,1\}^{\vnumunits}$
in one layer and \emph{$K$ binary} hidden units $\bh=\left[\hunit_{1},\hunit_{2},\text{\dots},\hunit_{K}\right]^{\top}\in\{0,1\}^{K}$
in the another layer. As an energy-based model, the RBM assigns the
energy function: $E\lrr{\visible,\bh;\paramset}=-\ba^{\top}\visible-\bb^{\top}\bh-\visible^{\top}\bW\bh$
where the parameter set $\paramset$ consists of visible biases $\ba=\left[a_{m}\right]_{M}\in\realset^{M}$,
hidden biases $\bb=\left[b_{k}\right]_{K}\in\realset^{K}$ and a weight
matrix $\bW=\left[w_{mk}\right]_{M\times K}\in\realset^{M\times K}$.
The element $w_{mk}$ represents the connection between the $m^{\text{th}}$
visible neuron and the $k^{\text{th}}$ hidden neuron.

Since each visible unit only connects to hidden units and vice versa,
the probability of a single unit being active in a layer only depends
on units in the other layer as below:
\begin{align}
p\left(\vunit_{m}=1\gv\bh;\paramset\right) & =\sigmoid\lrr{a_{m}+\boldsymbol{w}_{m\cdot}\bh}\label{eq:rbm_vis_cond}\\
p\left(h_{k}=1\mid\visible;\paramset\right) & =\sigmoid\lrr{b_{k}+\visible^{\top}\boldsymbol{w}_{\cdot k}}\label{eq:rbm_hid_cond}
\end{align}
This restriction on network architecture also introduces a nice property
of conditional independence between units at the same layer given
the another:\vspace{-0.6cm}

\begin{align}
p\left(\boldsymbol{h}\mid\visible;\paramset\right) & =\prod_{k=1}^{K}p\left(h_{k}\mid\visible;\paramset\right)\label{eq:rbm_hid_factorization}
\end{align}
\vspace{-0.8cm}

\begin{align}
p\left(\visible\mid\bh;\paramset\right) & =\prod_{m=1}^{M}p\left(\vunit_{m}\gv\bh;\paramset\right)\label{eq:rbm_vis_fractorization}
\end{align}
These factorizations also allow the data expectation in Eq.~\ref{eq:ebm_update}
to be computed analytically. Meanwhile, the model expectation still
remains intractable and requires an approximation, e.g., using Markov
Chain Monte Carlo (MCMC). However, sampling in the RBM can perform
efficiently using Gibbs sampling that alternatively draws the visible
and hidden samples from conditional distributions (Eqs. \ref{eq:rbm_hid_factorization}
and \ref{eq:rbm_vis_fractorization}) in one sampling step. The learning
can be accelerated with $d$-step Contrastive Divergence (denoted
$\text{CD}{}_{d}$) \cite{Hinton.Geoffrey_2002NeuralComp_CD}, which
considers the difference between the data distribution and the $d$-sampling
step distribution. $\text{CD}{}_{1}$ is widely-used because of its
high efficiency and small bias \cite{MiguelCarreiraPerpinan_etal_2005_CD}.
The following equations describe how $\text{CD}{}_{d}$ updates bias
and weight parameters using a minibatch of $N_{s}$ data samples. 

\vspace{-0.6cm}

\begin{align}
\vbiasunit_{m} & =\vbiasunit_{m}+\eta\frac{1}{N_{s}}\sum_{i=1}^{N_{s}}\lrr{\vunit_{m}^{\lrs i}-\hat{\vunit}_{m}^{\lra d}}\label{eq:rbm_grad_vis_bias}\\
b_{k} & =b_{k}+\eta\frac{1}{N_{s}}\sum_{i=1}^{N_{s}}\lrr{p\left(h_{k}=1\mid\visible^{\lrs i};\paramset\right)-\hat{\hunit}_{k}^{\lra d}}\label{eq:rbm_grad_hid_bias}\\
\w_{mk} & =\w_{mk}+\eta\frac{1}{N_{s}}\sum_{i=1}^{N_{s}}\lrr{\vunit_{m}^{\lrs i}p\left(h_{k}=1\mid\visible^{\lrs i};\paramset\right)-\hat{\vunit}_{m}^{\lra i}\hat{\hunit}_{k}^{\lra d}}\label{eq:rbm_grad_w}
\end{align}
wherein $\vunit_{m}^{\lrs i}$ is the $m^{\text{th}}$ element of
the $i^{\text{th}}$ training data vector whilst $\hat{\vunit}_{\cdot}^{\lra d}$
and $\hat{\hunit}_{\cdot}^{\lra d}$are visible and hidden samples
after $d$-sampling steps.

\subsection{Deep Boltzmann Machines\label{sub:Deep-Boltzmann-Machines}}

Deep Boltzmann Machine (DBM) \cite{Salakhutdinov.Ruslan_Hinton.Geoffrey_2009AISTATS}
is multilayer energy-based models, which enable to capture the data
distribution effectively and learn increasingly complicated representation
of the input. As a deep network, a \emph{binary} DBM consists of an
observed binary layer $\visible$ of $\vnumunits$ units and many
\emph{binary} hidden layers. For simplicity, we just consider a DBM
with two hidden layers $\hidden=\{\hiddenone,\hiddentwo\}$ of $\hnumunitsone$
and $\hnumunitstwo$ units respectively. Similar to RBMs, the DBM
defines a visible bias vector $\vbias$ and a hidden bias vector $\hbias^{\lrr l}$
for the hidden layer $\hidden^{\lrr l}$. Two adjacent layers communicate
with each other through a full connection including a visible-to-hidden
matrix $\Wone$ and a hidden-to-hidden matrix $\Wtwo$. The energy
of joint configuration $\lrr{\visible,\hidden}$ with respect to the
parameter set $\paramset=\{\vbias,\hbiasone,\hbiastwo,\Wone,\Wtwo\}$
is represented as: \vspace{-0.5cm}

\begin{eqnarray*}
E\lrr{\visible,\hidden;\paramset} & = & -\vbias^{\top}\visible-\hbiasoneT\hiddenone-\hbiastwoT\hiddentwo\\
 &  & -\visible^{\top}\Wone\hiddenone-\hiddenoneT\Wtwo\hiddentwo
\end{eqnarray*}

Like RBMs, there is a requirement on no connection between units in
the same layer and then the conditional probability of a unit to
be $1$ given the upper and the lower layers is as follows:

\vspace{-0.6cm}

\begin{align}
p\lrr{v_{m}=1|\hiddenone;\paramset} & =\sigmoid\lrr{a_{m}+\boldsymbol{w}_{m\cdot}^{\lrr 1}\hiddenone}\label{eq:dbm_cond_prob_v}\\
p\lrr{\hunit_{n}^{\lrr 1}=1|\visible,\hiddentwo;\paramset} & =\sigmoid\lrr{\hbiasunit_{n}^{\lrr 1}+\visible^{\top}\boldsymbol{w}_{\cdot n}^{\lrr 1}+\boldsymbol{w}_{n\cdot}^{\lrr 2}\hiddentwo}\label{eq:dbm_cond_prob_h1}\\
p\lrr{\hunit_{n}^{\lrr 2}=1|\hiddenone;\paramset} & =\sigmoid\lrr{\hbiasunit_{n}^{\lrr 2}+\hiddenoneT\boldsymbol{w}_{\cdot n}^{\lrr 2}}\label{eq:dbm_cond_prob_h2}
\end{align}

To train DBM, we need to deal with both intractable expectations.
The data expectation is usually approximated by its lower bound that
is computed via a factorial variational distribution:

\vspace{-0.6cm}

\begin{equation}
q\lrr{\hiddenone,\hiddentwo;\varparamset}=\prod_{l=1}^{2}\prod_{i=1}^{\hnumunits_{l}}\lrr{\varmean_{i}^{\lrr l}}^{\hunit_{i}^{\lrr l}}\lrr{1-\varmean_{i}^{\lrr l}}^{1-\hunit_{i}^{\lrr l}}\label{eq:dbm_variational_ditribution}
\end{equation}
wherein $\varmean$ are variational parameters and learned by updating
iteratively the fixed-point equations below: 

\vspace{-0.5cm}

\begin{eqnarray}
\varmean_{n}^{\lrr 1} & = & \sigmoid\lrr{\hbiasunit_{n}^{\lrr 1}+\visible^{\top}\boldsymbol{w}_{\cdot n}^{\lrr 1}+\boldsymbol{w}_{n\cdot}^{\lrr 2}\boldsymbol{\varmean}^{\lrr 2}}\label{eq:dbm_variational_param_h1}\\
\varmean_{n}^{\lrr 2} & = & \sigmoid\lrr{\hbiasunit_{n}^{\lrr 2}+\boldsymbol{\varmean}^{\lrr 1\top}\boldsymbol{w}_{\cdot n}^{\lrr 2}}\label{eq:dbm_variational_param_h2}
\end{eqnarray}
For model expectation, the conditional dependence of intra-layer units
again allows to employ Gibbs sampling alternatively between the odd
and even layers. The alternative sampling strategy is used in the
popular training method of Persistent Contrastive Divergence (PCD)
\cite{Tieleman.Tijmen_2008ICML_PCD} that maintains several persistent
Gibbs chains to provide the model samples for training. In every
iteration, given a batch of $N_{s}$ data points, its mean-field vectors
and samples on $N_{\text{c}}$ Gibbs chains are computed and the model
parameters are updated using the following equations: \vspace{-0.4cm}

\begin{align}
\Delta\vbiasunit_{m} & =\eta\lrr{\sum_{i=1}^{N_{s}}\frac{\vunit_{m}^{\lrs i}}{N_{s}}-\sum_{i=1}^{N_{\text{c}}}\frac{\hat{\vunit}_{m}^{\lra i}}{N_{\text{c}}}}\label{eq:dbm_update_visible_bias}\\
\Delta b_{n}^{\lrr l} & =\eta\lrr{\sum_{i=1}^{N_{s}}\frac{\varmean_{n}^{\lrr l\lrs i}}{N_{s}}-\sum_{i=1}^{N_{\text{c}}}\frac{\hat{\hunit}_{n}^{\lrr l\lra i}}{N_{\text{c}}}}\label{eq:dbm_update_hidden_bias}\\
\Delta\w_{mn}^{\lrr 1} & =\eta\lrr{\sum_{i=1}^{N_{s}}\frac{\vunit_{m}^{\lrs i}\varmean_{n}^{\lrr 1\lrs i}}{N_{s}}-\sum_{i=1}^{N_{\text{c}}}\frac{\hat{\vunit}_{m}^{\lra i}\hat{\hunit}_{n}^{\lrr 1\lra i}}{N_{\text{c}}}}\label{eq:dbm_update_W1}\\
\Delta\w_{mn}^{\lrr 2} & =\eta\lrr{\sum_{i=1}^{N_{s}}\frac{\varmean_{m}^{\lrr 1\lrs i}\varmean_{n}^{\lrr 2\lrs i}}{N_{s}}-\sum_{i=1}^{N_{\text{c}}}\frac{\hat{\hunit}_{m}^{\lrr 1\lra i}\hat{\hunit}_{n}^{\lrr 2\lra i}}{N_{\text{c}}}}\label{eq:dbm_update_W2}
\end{align}
wherein $\visible^{\lrs i}$ and $\varmeanlayer^{\lrr l\lrs i}$ are
the $i^{\text{th}}$ data point and its corresponding mean-field vector
whilst $\hat{\visible}^{\lra i}$and $\hat{\hidden}^{\lrr l\lra i}$
are layer states on the $i^{\text{th}}$ Gibbs chain.

In addition to variational approximation and PCD, a greedy layer-wise
pretraining \cite{Salakhutdinov.Ruslan_Hinton.Geoffrey_2009AISTATS,Salakhutdinov.Ruslan_Hinton.Geoffrey_2012_DBM}
is necessary to guarantee the best performance of the trained models.

\subsection{Data reconstruction}

Once the RBM or the DBM has been learned, it is able to reconstruct
any given data $\visible$. In particular, we can project the data
$\visible$ into the space of the first hidden layer for the new representation
$\hidden_{r}=[\tilde{\hunit}_{1},\tilde{\hunit}_{2},...,\tilde{\hunit}_{\hnumunitsone}]^{\top}$
by computing the posterior $\tilde{\hunit}_{n}=p\left(\hunit_{n}=1\gv\visible;\paramset\right)$
in RBMs or running mean-field iterations to estimate $\tilde{\hunit}_{n}=\varmean_{n}^{\lrr 1}$
in DBMs. Next, projecting back this representation into the input
space forms the reconstructed output $\visible_{r}=\left[\tilde{\vunit}_{1},\tilde{\vunit}_{2},...,\tilde{\vunit}_{M}\right]^{\top}$,
where $\tilde{\vunit}_{m}$ is shorthand for $\tilde{\vunit}_{m}=p\left(\vunit_{m}=1\gv\bh_{r};\paramset\right)$.
Finally, the reconstruction error is simply the difference between
two vectors $\visible$ and $\visible_{r}$, where we prefer the Euclidean
distance due to its popularity. If $\visible$ belongs to the group
of normal events, which the model is learned well, the reconstructed
output is almost similar to $\visible$ in terms of low reconstruction
error. By contrast, an abnormal event usually causes a high error.
For this reason, we use the reconstruction quality of models as a
signal to identify anomalous events.

\section{Framework\label{sec:Framework}}

\begin{figure}[t]
\begin{centering}
\emph{\includegraphics[width=1\columnwidth]{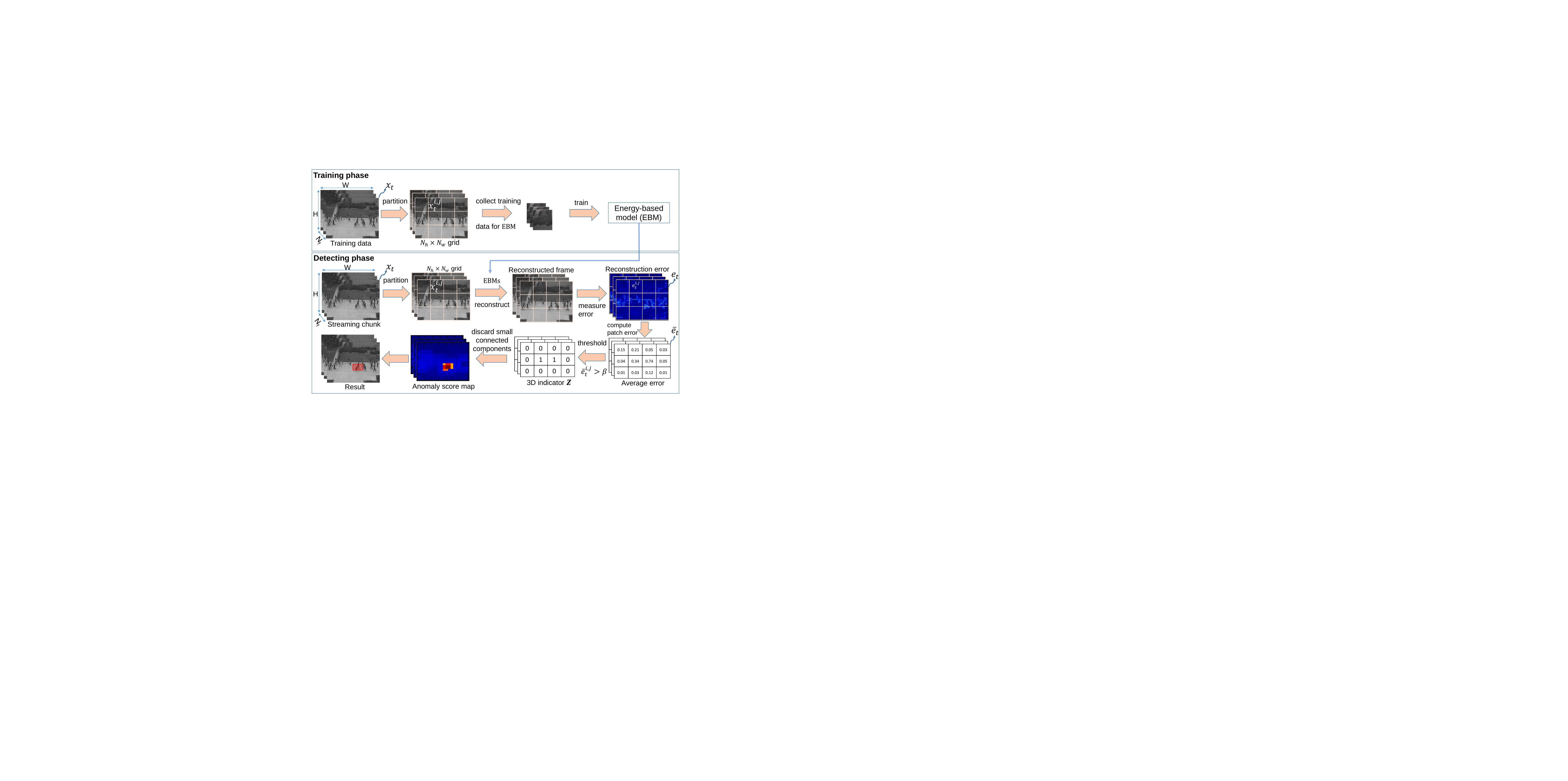}}\vspace{-2mm}

\par\end{centering}

\centering{}\caption{The overview of our proposed framework.\label{figFramework}}
\vspace{-2mm}
\end{figure}

\begin{figure}[t]
\begin{centering}
\emph{\includegraphics[width=1\columnwidth]{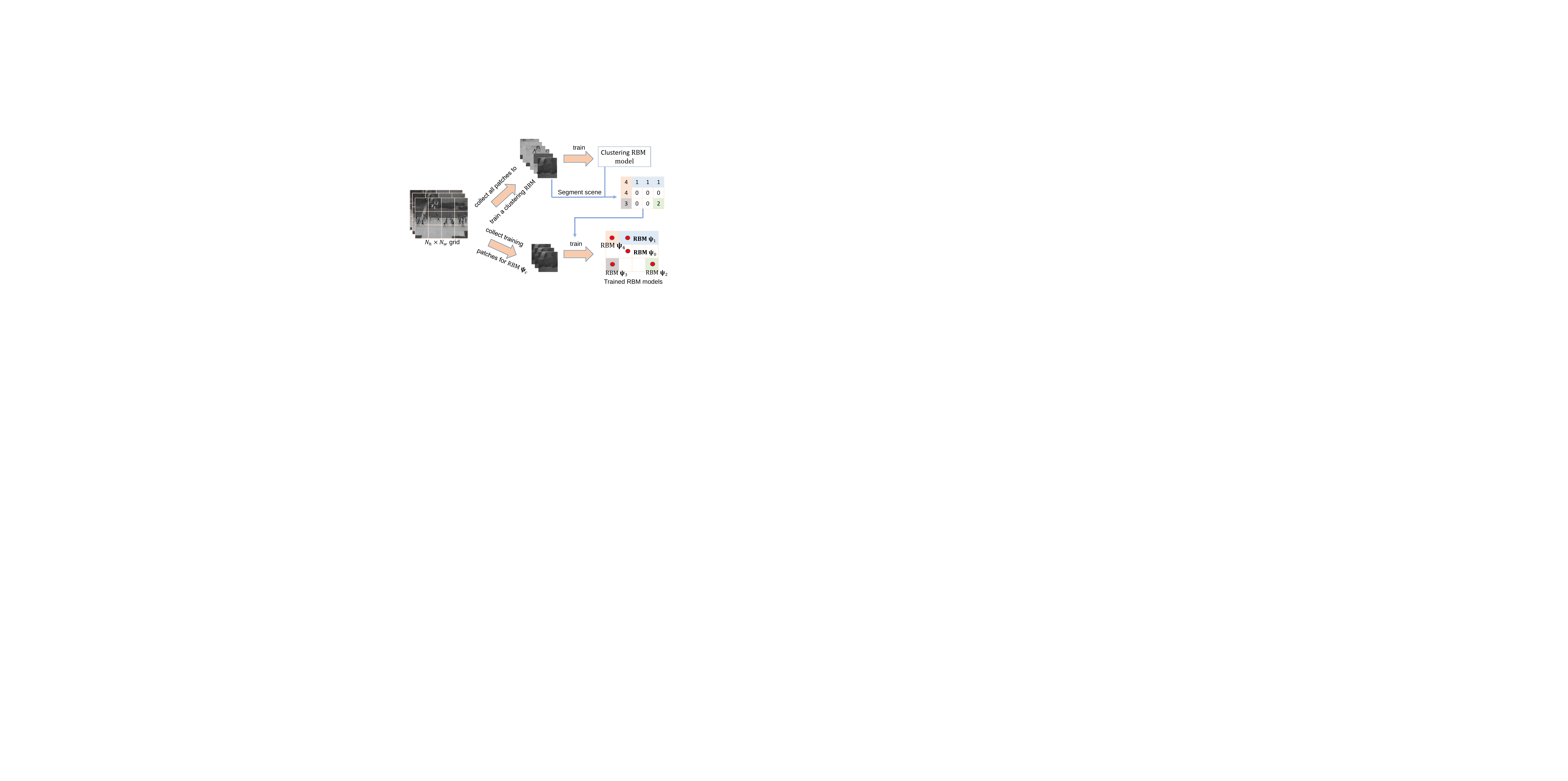}}
\par\end{centering}

\begin{centering}
\vspace{-2mm}

\par\end{centering}

\caption{Scheme to train $\protect\modelRBM$ with a clustering RBM and region
RBMs.\label{figEAD-RBM}}
\vspace{-5mm}
\end{figure}

\begin{figure}[t]
\begin{centering}
\vspace{2mm}
\emph{\includegraphics[width=1\columnwidth]{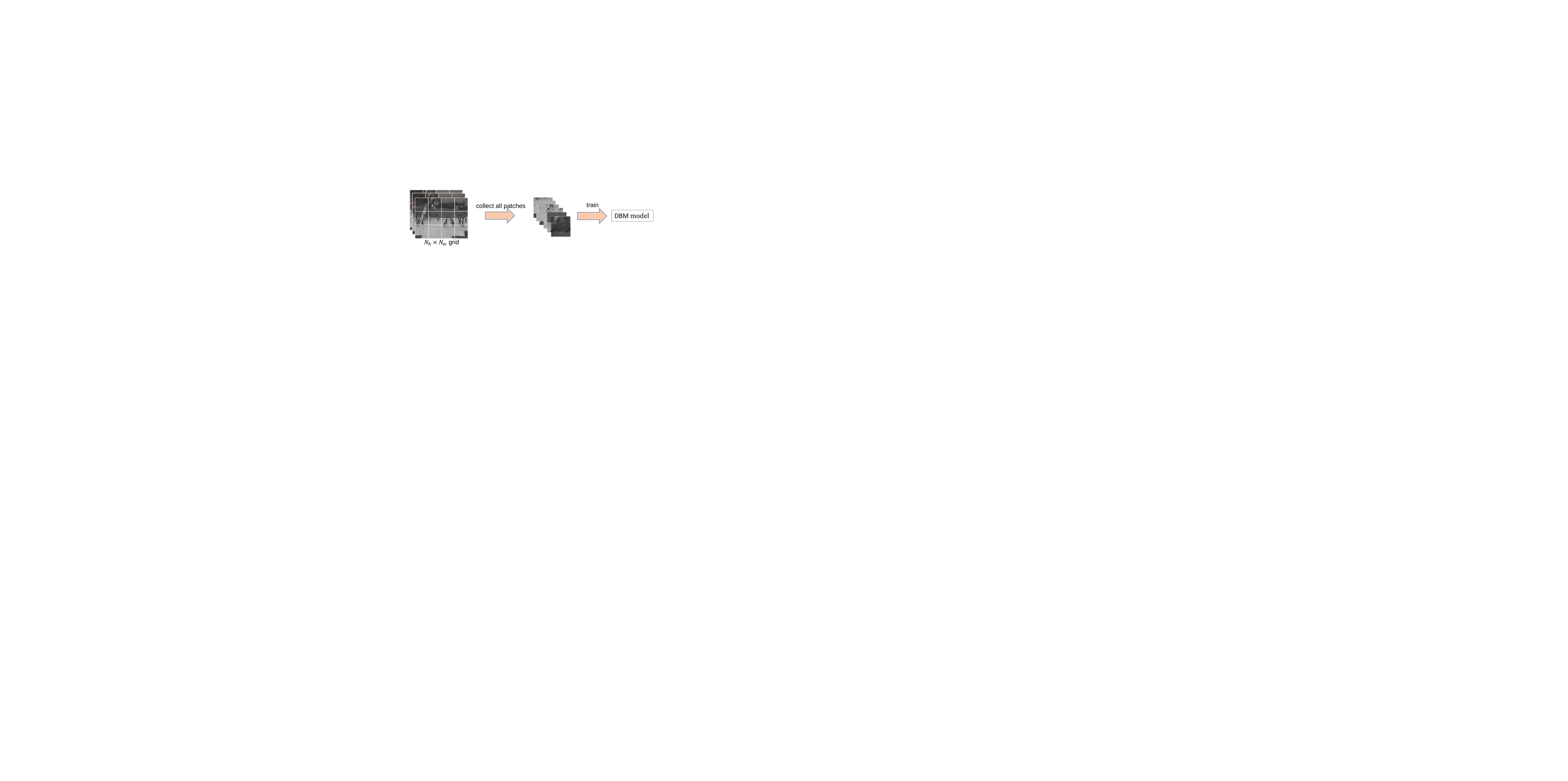}}
\par\end{centering}

\begin{centering}
\vspace{-2mm}

\par\end{centering}

\caption{Scheme to train $\protect\modelDBM$ with a single DBM.\label{figEAD-DBM}}
\vspace{-5mm}
\end{figure}

This section describes our proposed framework of Energy-based Anomaly
Detection ($\modelEAD$) to localize anomaly events in videos. In
general, an $\modelEAD$ system is a two-phase pipeline of a training
phase and a detection phase as demonstrated in Fig. \ref{figFramework}.
The training phase consists of three steps: (i) treating videos as
a collection of images and splitting frames into a grid of patches;
(ii) gathering patches and vectorizing them; (iii) training EBM models.
In detection phase, the $\modelEAD$ system: (i) decomposes videos
into patches; (ii) feeds patches into the trained $\modelEBM$s for
reconstructed frames and reconstruction error maps; (iii) selects
regions with the high probability of being abnormal by thresholding
the error maps and represents surviving regions as graphical connected
components and then filters out the small anomaly objects corresponding
to small-sized components; and finally (iv) updates the $\modelEBM$s
incrementally with video stream data. In what follows, we explain
these phases in more details.

\subsection{Training phase }

Suppose that we denote a video of $N_{f}$ frames as $D=\{\visible_{t}\in\mathbb{R}^{H\times W}\}_{t=1}^{N_{f}}$,
where $H$ and $W$ are the frame size in pixel. Theoretically, we
can vectorize the video frames and train the models on data vectors
of $H\times W$ dimensions. However, $H\times W$ is extremely large
in real-life videos, e.g., hundreds of thousand pixels, and hence
it is infeasible to train $\modelEBM$s in high-dimensional image
space. This is because the high-dimensional input requires more complex
models with an extremely large number of parameters (i.e., millions).
This makes the parameter learning more difficult and less robust since
it is hard to control the bounding of hidden activation values. Thus
the hidden posteriors are easily collapsed into either zeros or ones,
and no more learning occurs.

Another solution is to do dimensionality reduction, which projects
the frames in the high dimensional input space into a subspace with
lesser dimensions. But employing this solution agrees a sacrifice
in terms of losing rich source of information in original images.
To preserve the full information as well as reduce the data dimensionality,
we choose to apply $\modelEBM$s to image patches instead of the whole
frames. In other words, we divide every frame $\visible_{t}$ into
a grid of $N_{h}\times N_{w}$ patches $\visible_{t}=\{\visible_{t}^{i,j}\gv1\leq i\leq N_{h},1\leq j\leq N_{w}\}$
using the patch size of $h\times w$. These patches are flattened
into vectors and gathered into a data collection to train models.

\subsubsection{RBM-based framework}

Once patch data is available, we have two possible ways to train the
models: a) learn one individual RBM on patches at the same location
($i,j$) or b) learn only one RBM on all patches in the videos. The
first choice results in the excessive number of models, e.g., approximate
$400$ models to work on  the $240\times360$ video resolution and
the non-overlapping patch size of $12\times18$ pixels,  rendering
very high computational complexity and memory demand. Meanwhile,
the single model approach ignores the location information of events
in videos. An example is the video scene of vehicles on a street and
pedestrians on a footpath. Such model cannot recognize the emergency
cases when a car mounts the footpath or people suddenly cross the
street without zebra-crossings. 

Our solution is to reduce the number of models and preserve the location
infomration by grouping similar patches at some locations and training
one model for each group. This proposal is based on our observation
that image patches of the same terrains, buildings or background regions
(e.g., pathways, grass, streets, walls, sky or water) usually share
the same appearance and texture. Therefore, using many models to represent
the similar patches is redundant and they can be replaced by one shared
model. To that end, we firstly cluster the video scene into similar
regions by training a RBM with a few hidden units (i.e., $\hnumunits=4$)
on all patches. To assign a cluster to a patch $\visible_{t}^{i,j}$,
we compute the hidden representation $\hidden_{r}$ of the patch and
binarize it to obtain the binary vector $\tilde{\bh}=\left[\mathbb{I}\left(\tilde{h}_{1}>0.5\right),...,\mathbb{I}\left(\tilde{h}_{K}>0.5\right)\right]$
where $\mathbb{I}\left(\cdot\right)$ is the indicator function. The
cluster label of $\visible_{t}^{i,j}$ is the decimal value of the
binary vector, e.g., 0101 converted to 5. Afterwards, we compute the
region label $c^{i,j}$ at location ($i,j$) by voting the labels
of patches at ($i,j$) over the video frames. As a result, the similar
regions of the same footpaths, walls or streets are assigned to the
same label numbers and the video scene is segmented into $C$ similar
regions. For each region $c$, we train a RBM parameter set $\paramset_{c}$
on all patches belonging to the region. After training phase, we comes
up with an $\modelRBM$ system with one clustering RBM and $C$ region
RBMs. Fig. \ref{figEAD-RBM} summarizes the training procedure of
our $\modelRBM$.

\subsubsection{DBM-based framework}

Although $\modelRBM$ can reduce the number of models dramatically,
$\modelRBM$ requires to train $C+1$ models, e.g., $C=16$ if $\hnumunits=4$.
This training procedure (Fig. \ref{figEAD-RBM}) is still complicated.
Further improvement can be done by extending the $\modelRBM$ using
DBMs whose multilayer structure offers more powerful capacity than
the shallow structure of RBMs. In particular, one hidden layer in
RBMs offers either clustering or reconstruction capacity per network
whilst the multilayer networks of DBMs allow to perform multitasking
in the same structure. In this work, we propose to integrate a DBM
as demonstrated in Fig.~\ref{figEAD-DBM} into $\modelEAD$ to detect
abnormality. This network consists of two hidden layers $\hiddenone$
and $\hiddentwo$ and two visible layers $\visible^{\lrr 1}$ and
$\visible^{\lrr 2}$ at its ends. The data is always fed into both
$\visible^{\lrr 1}$ and $\visible^{\lrr 2}$ simultaneously. The
first hidden layer has $\hnumunits$ units and it has responsibility
to do a clustering task. Meanwhile, the second hidden layer has a
lot of units to obtain good reconstruction capacity. These layers
directly communicate with data to guarantee that the learned model
can produce good examplars and reconstruction of the data. Using
the proposed architecture, one DBM has the equivalent power to $C+1$
RBMs in $\modelRBM$ system. Therefore, it is an appealing alternative
to both clustering RBM and region RBMs in $\modelRBM$. Furthermore,
we only need to train one DBM, rendering a significant improvement
in the number of trained models. 

\begin{figure}[t]
\begin{centering}
\includegraphics[width=0.6\columnwidth]{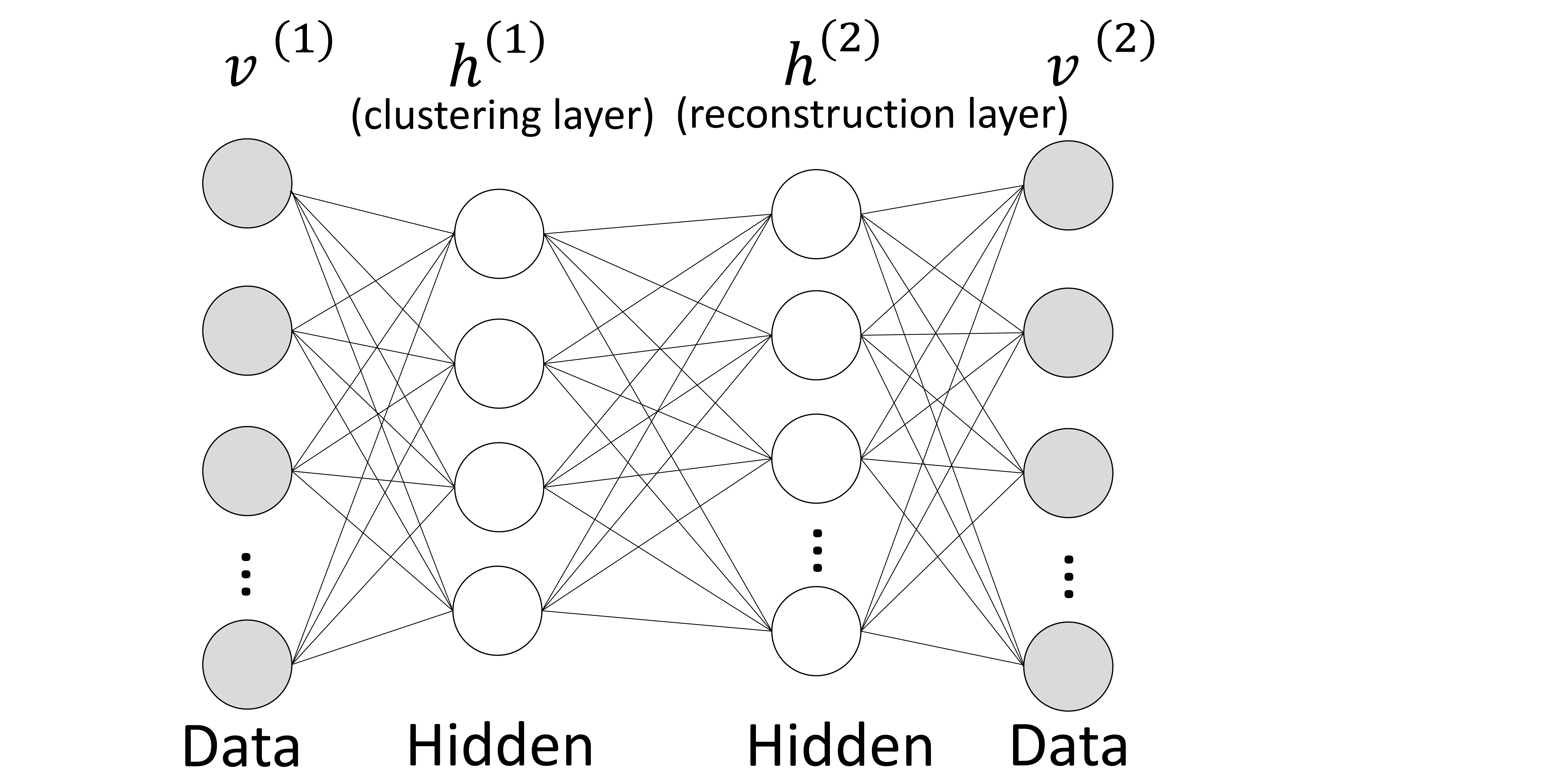}
\par\end{centering}

\caption{The architecture of our clustering and reconstruction DBM.}
\label{figTwoEndDBM}
\end{figure}

To train this DBM, we employ the PCD procedure, the variational approximation
and the layer-wise pretraining step as described in Sec. \ref{sub:Deep-Boltzmann-Machines}
using the equations in Table \ref{tablemultitaskdbm}. In addition,
to improve the reconstruction quality of the trained model, we use
conditional probabilities (Eqs. \ref{eq:multitaskdbm_cond_prob_v1}-\ref{eq:multitaskdbm_cond_prob_v2}
in Table \ref{tablemultitaskdbm}) as states of units rather than
sampling them from these probabilities. This ensures to diversify
the states of neurons and strengthen the reconstruction capacity of
the network. But it is noteworthy that an exception is units on the
first hidden layer $\hiddenone$ whose states are still binary. This
is because $\hiddenone$ has responsibility to represent data clusters
and therefore it should have limited states. A DBM's variant that
is close to our architecture is Multimodal DBMs \cite{Srivastava.Nitis_Salakhutdinov.Ruslan_2014JMLR_BinaryGaussianDBM}.
In that study, the different types of data, e.g., images and texts,
are attached into two ends of the network in order to model the joint
representation across data types. By contrast, our architecture is
designed  to do multitasks. To the best of our knowledge, our proposed
network of both reconstruction and clustering capacities is distinct
from other DBM's studies in the literature. 

\begin{table*}[th]
\begin{centering}
\resizebox{0.9\textwidth}{!}{%%
\begin{tabular}{>{\raggedright}p{0.57\textwidth}|>{\raggedright}p{0.4\textwidth}}
\textbf{Energy function:}

\begin{minipage}[t]{1.17\columnwidth}%
\vspace{-0.3cm}
\begin{eqnarray}
E\lrr{\visible,\hidden;\paramset} & = & -\vbias^{\lrr 1\top}\visible^{\lrr 1}-\vbias^{\lrr 2\top}\visible^{\lrr 2}-\hbiasoneT\hiddenone-\hbiastwoT\hiddentwo\nonumber \\
 &  & -\visible^{\lrr 1\top}\Wone\hiddenone-\hiddenoneT\Wtwo\hiddentwo-\hiddentwoT\Wthree\visible^{\lrr 2}\label{eq:multitaskdbm_energy_function}
\end{eqnarray}

\vspace{-0.1cm}
\end{minipage} & \multirow{3}{0.4\textwidth}{\textbf{Parameter update equations:}\\
\begin{minipage}[t]{1\columnwidth}%
\vspace{-0.2cm}
\begin{eqnarray}
\Delta\vbiasunit_{m}^{\lrr l} & = & \eta\lrr{\sum_{i=1}^{N_{s}}\frac{\vunit_{m}^{\lrr l\lrs i}}{N_{s}}-\sum_{i=1}^{N_{\text{c}}}\frac{\hat{\vunit}_{m}^{\lrr l\lra i}}{N_{\text{c}}}}\label{eq:multitaskdbm_update_visible_bias}\\
\Delta b_{n}^{\lrr l} & = & \eta\lrr{\sum_{i=1}^{N_{s}}\frac{\varmean_{n}^{\lrr l\lrs i}}{N_{s}}-\sum_{i=1}^{N_{\text{c}}}\frac{\hat{\hunit}_{n}^{\lrr l\lra i}}{N_{\text{c}}}}\label{eq:multitaskdbm_update_hidden_bias}\\
\Delta\w_{mn}^{\lrr 1} & = & \eta\lrr{\sum_{i=1}^{N_{s}}\frac{\vunit_{m}^{\lrr 1\lrs i}\varmean_{n}^{\lrr 1\lrs i}}{N_{s}}-\sum_{i=1}^{N_{\text{c}}}\frac{\hat{\vunit}_{m}^{\lrr 1\lra i}\hat{\hunit}_{n}^{\lrr 1\lra i}}{N_{\text{c}}}}\label{eq:multitaskdbm_update_W1}\\
\Delta\w_{mn}^{\lrr 2} & = & \eta\lrr{\sum_{i=1}^{N_{s}}\frac{\varmean_{m}^{\lrr 1\lrs i}\varmean_{n}^{\lrr 2\lrs i}}{N_{s}}-\sum_{i=1}^{N_{\text{c}}}\frac{\hat{\hunit}_{m}^{\lrr 1\lra i}\hat{\hunit}_{n}^{\lrr 2\lra i}}{N_{\text{c}}}}\label{eq:multitaskdbm_update_W2}\\
\Delta\w_{nm}^{\lrr 3} & = & \eta\lrr{\sum_{i=1}^{N_{s}}\frac{\varmean_{n}^{\lrr 2\lrs i}\vunit_{m}^{\lrr 2\lrs i}}{N_{s}}-\sum_{i=1}^{N_{\text{c}}}\frac{\hat{\hunit}_{n}^{\lrr 2\lra i}\hat{\vunit}_{m}^{\lrr 2\lra i}}{N_{\text{c}}}}\label{eq:multitaskdbm_update_W3}
\end{eqnarray}
\end{minipage}}\tabularnewline
\cline{1-1} 
\vspace{-0.1cm}
\textbf{Conditional probabilities:}

\begin{minipage}[t]{1.1\columnwidth}%
\vspace{-0.3cm}
\begin{eqnarray}
p\lrr{v_{m}^{\lrr 1}=1|\hiddenone;\paramset}\text{\,\,\,\,\,\,\,\,\,\,\,\,\,} & = & \sigmoid\lrr{a_{m}^{\lrr 1}+\boldsymbol{w}_{m\cdot}^{\lrr 1}\hiddenone}\label{eq:multitaskdbm_cond_prob_v1}\\
p\lrr{\hunit_{n}^{\lrr 1}=1|\visible^{\lrr 1},\hiddentwo;\paramset} & = & \sigmoid\lrr{\hbiasunit_{n}^{\lrr 1}+\visible^{\lrr 1\top}\boldsymbol{w}_{\cdot n}^{\lrr 1}+\boldsymbol{w}_{n\cdot}^{\lrr 2}\hiddentwo}\label{eq:multitaskdbm_cond_prob_h1}\\
p\lrr{\hunit_{n}^{\lrr 2}=1|\hiddenone,\visible^{\lrr 2};\paramset} & = & \sigmoid\lrr{\hbiasunit_{n}^{\lrr 2}+\hiddenoneT\boldsymbol{w}_{\cdot n}^{\lrr 2}+\boldsymbol{w}_{n\cdot}^{\lrr 3}\visible^{\lrr 2}}\label{eq:multitaskdbm_cond_prob_h2}\\
p\lrr{v_{m}^{\lrr 2}=1|\hiddentwo;\paramset}\text{\,\,\,\,\,\,\,\,\,\,\, } & = & \sigmoid\lrr{a_{m}^{\lrr 2}+\hiddentwoT\boldsymbol{w}_{\cdot m}^{\lrr 3}}\label{eq:multitaskdbm_cond_prob_v2}
\end{eqnarray}

\vspace{-0.1cm}
\end{minipage} & \tabularnewline
\cline{1-1} 
\vspace{-0.1cm}
\textbf{Mean-field update equations:}

\begin{minipage}[t]{1.1\columnwidth}%
\vspace{-0.3cm}
\begin{eqnarray}
\varmean_{n}^{\lrr 1} & = & \sigmoid\lrr{\hbiasunit_{n}^{\lrr 1}+\visible^{\lrr 1\top}\boldsymbol{w}_{\cdot n}^{\lrr 1}+\boldsymbol{w}_{n\cdot}^{\lrr 2}\boldsymbol{\varmean}^{\lrr 2}}\label{eq:multitaskdbm_variational_param_h1}\\
\varmean_{n}^{\lrr 2} & = & \sigmoid\lrr{\hbiasunit_{n}^{\lrr 2}+\boldsymbol{\varmean}^{\lrr 1\top}\boldsymbol{w}_{\cdot n}^{\lrr 2}+\boldsymbol{w}_{n\cdot}^{\lrr 3}\visible^{\lrr 2}}\label{eq:multitaskdbm_variational_param_h2}
\end{eqnarray}
\end{minipage} & \tabularnewline
\end{tabular}}
\par\end{centering}

\caption{Equations of a clustering reconstruction DBM. }
\vspace{-0.5cm}
\label{tablemultitaskdbm}
\end{table*}

\subsection{Detection phase}

Once $\modelRBM$ or $\modelDBM$ has been learned from training data,
we can use it to detect anomalous events in testing videos. The Alg.
\ref{algDetectionPhase} shows the pseudocode of this phase that can
be summarized into three main steps of: a) reconstructing frames and
computing reconstruction errors; b) localizing the anomaly events
and c) updating the $\modelEBM$s incrementally. In what follows,
we introduce these steps in more details.

\setlength{\textfloatsep}{0.2cm}
%\begin{wrapfigure*}[25]{r}{1.0\columnwidth}     
%\begin{minipage}{0.95\columnwidth}

\begin{algorithm}[t] 
\begin{algorithmic}[1]
\small
\Require{ Chunk $\left\{ \visible_{t}\right\} _{t=1}^{L}$, models $\paramset$, thresholds $\beta$ and $\gamma$}
\Ensure{Detection $\boldsymbol{Z}$, score $\left\{ \bar{e}_{t}^{i,j}\right\} $}
\For {$t\leftarrow1,\ldots,L$}
\For {$\visible_{t}^{i,j}\in\visible_{t}$}
\State{$\visible_{\text{r},t}^{i,j}\leftarrow$reconstruct($\visible_{t}^{i,j}$,$\paramset$)}
\State{$\boldsymbol{e}_{t}^{i,j}\leftarrow|\visible_{t}^{i,j}-\visible_{\text{r},t}^{i,j}|$}
\State{$\bar{e}_{t}^{i,j}\leftarrow\frac{1}{h\times w}\left\Vert \boldsymbol{e}_{t}^{i,j}\right\Vert _{2}$}
\If{$\bar{e}_{t}^{i,j}\geq\beta$} 
\For {$p\in\visible_{t}^{i,j}$}
\State{$\bZ(p)\leftarrow1$}
\EndFor
\Else
\For {$p\in\visible_{t}^{i,j}$}
\State{$\bZ(p)\leftarrow0$}
\EndFor
\EndIf
\EndFor
\State{//Online version only}
\For {$c\leftarrow1,\ldots,C$}
\State{$\boldsymbol{X}_{t}^{c}\leftarrow\left\{ \visible_{t}^{i,j}\mid c^{i,j}=c\right\} $}
\State{$\paramset_{c}\leftarrow\text{{updateRBM}}(\boldsymbol{X}_{t}^{c},\paramset)$}
\EndFor
\EndFor
\State{$\boldsymbol{Z}\leftarrow$remove\_small\_components($\boldsymbol{Z},$$\gamma$)}
\normalsize
\end{algorithmic}
\caption{Detection with \modelEAD} 
\label{algDetectionPhase} 
\end{algorithm}

%\end{minipage}   
%\end{wrapfigure*}

At first, the video stream is split into chunks of $L$ non-overlapping
frames $\left\{ \visible_{t}\right\} _{t=1}^{L}$ which next are partitioned
into patches $\visible_{t}^{i,j}$ as the training phase. By feeding
these patches into the learned $\modelEAD$s, we obtain the reconstructed
patches $\visible_{\text{r,}t}^{i,j}$ and the reconstruction errors
$\be_{t}^{i,j}=|\visible_{t}^{i,j}-\visible_{\text{r},t}^{i,j}|$.
One can use these errors to identify anomaly pixels by comparing them
with a given threshold. However, these pixel-level reconstruction
errors are not reliable enough because they are sensitive to noise.
As a result, this approach may produce many false alarms when normal
pixels are reconstructed with high errors, and may fail to cover the
entire abnormal objects in such a case that they are fragmented into
isolated high error parts. Our solution is to use the patch average
error $\bar{e}_{t}^{i,j}=||\boldsymbol{e}_{t}^{i,j}||_{2}/\left(h\times w\right)$
rather than the pixel errors. If $\bar{e}_{t}^{i,j}\geq\beta$, all
pixels in the corresponding patch are assigned to be abnormal. 

After abnormal pixels in patches are detected in each frame, we concatenate
$L$ contiguous detection maps to obtain a 3D binary hyperrectangle
$\bZ\in\left\{ 0,1\right\} ^{L\times H\times W}$ wherein $z_{i,j,k}=1$
indicates an abnormal voxel and otherwise $0$ is a normal one. Throughout
the experiments, we observe that although most of the abnormal voxels
in $\bZ$ are correct, there are a few groups of abnormal voxels that
are false detections because of noise. To filter out these voxels,
we firstly build a sparse graph whose vertices are abnormal voxels
$z_{i,j,k}=1$ and edges are connections between two vertices $z_{i,j,k}$
and $z_{i+m,j+n,k+t}$ satisfying $m,n,t\in\left\{ -1,0,1\right\} $
and $\left|m\right|+\left|n\right|+\left|t\right|>0$. Then, we apply
a connected component algorithm to this graph and remove noisy components
that are defined to span less than $\gamma$ contiguous frames. 
The average errors $\bar{e}_{t}^{i,j}$ after this component filtering
step can be used as a final anomaly score. 

One problem is that objects can appear at different sizes and scales
in videos. To tackle this problem, we independently employ the detection
procedure above in the same videos at different scales. This would
help the patch partially or entirely cover objects at certain scales.
In particular, we rescale the original video into different resolutions,
and then compute the corresponding final anomaly maps and the binary
3D indicator tensors $\bZ$. The final anomaly maps at these scales
are aggregated into one map using a max-operation in $\modelRBM$
and a mean-operation in $\modelDBM$. The mean-operation is used in
$\modelDBM$ is because we observe that DBMs at the finer resolutions
usually cover more patches and they tend to over-detect whilst models
at the coarser resolutions prefer under-detecting. Averaging maps
at different scales can address these issues and produce better results.
For $\modelRBM$, since region RBMs frequently work in image segments
and are rarely affected by scales, we can pick up the best maps over
resolution. Likewise, the binary indicator tensors $\bZ$ are also
combined into one tensor using a binary OR-operation before proceeding
the connected component filtering step. In this work, we use overlapping
patches for better detection accuracy. The pixels in overlapping regions
are averaged when combining maps and indicator tensors at different
scales.

\subsubsection{Incremental detection}

In the scenario of data streaming where videos come on frame by frame,
the scene frequently changes over time and the current frame is significantly
different from those are used to train models. As a result, the models
become out of date and consider all regions as abnormalities. To handle
this problem, we let the models be updated with new frames. More specifically,
for every oncoming frame $t$, we use all patches with region label
$c$ to update the RBM $\paramset_{c}$ in $\modelRBM$. The updating
procedure is exactly the same as parameter updates (Eqs. \ref{eq:rbm_grad_vis_bias}-\ref{eq:rbm_grad_w})
in training phase using gradient ascent and $20$ epochs. Here we
use several epochs to ensure that the information of new data are
sufficiently captured by the models. 

\setlength{\textfloatsep}{0.2cm}
%\begin{wrapfigure}[20]{r}{1.0\columnwidth}     
%\begin{minipage}{0.95\columnwidth}

\begin{algorithm}[t] 
\begin{algorithmic}[1]
\small
\Require{ Training data $\mathcal{D}=\left\{ \visible^{\lrs i}\right\} _{i=1}^{N}$, original RBM of $M$ visible and $K$ hidden units and weights $\bW$, \# selected hidden units $K^{\prime}$}
\Ensure{New RBM with weights $\bW^{\prime}$}

\For {$i\leftarrow1,\ldots,N$}
\For {$n\leftarrow1,\ldots,K$}
\State{$\tilde{\hunit}_{n}^{\lrs i}\leftarrow p\left(\hunit_{n}=1\gv\visible^{\lrs i}\right)$}

\EndFor

\EndFor
\For {$n\leftarrow1,\ldots,K$}
\State{$\alpha_{n}\leftarrow \sum_{i=1}^N\sum_{m=1}^M\frac{\lrv {w_{mn}\tilde{\hunit}_{n}^{\lrs i}}}{NM}$}
\EndFor
\State{$j_{1},...,j_{K^{\prime}}\leftarrow \text{index\_of\_top\_max\_of}{\lrs {\alpha_n}}$}

\State{$\bW^{\prime}\leftarrow \lrs{ \bw_{\cdot j_1},...,\bw_{\cdot j_{K^{\prime}}}}$}
\normalsize
\end{algorithmic}
\caption{Reduce RBM hidden units} 
\label{algRBMReduction} 

\end{algorithm}
%\end{minipage}   
%\end{wrapfigure}

For $\modelDBM$, updating one DBM model for the whole scene is ineffective.
The reason is that, in a streaming scenario, a good online system
should have a capacity of efficiently adapting itself to the rapid
changes of scenes using limited data of the current frames. These
changes, e.g., new pedestrians,  occur gradually in some image patches
among a large number of static background patches, e.g., footpaths
or grass. However, since a single DBM has to cover the whole scene,
it is usually distracted by these background patches during its updating
and becomes insensitive to such local changes. As a result, there
is an insufficient difference in detection quality between updated
and non-updated DBM models. Our solution is to build region DBMs,
each of which has responsibility for monitoring patches in the corresponding
region. Because each DBM observes a smaller area, it can instantly
recognize the changes in that area. These region DBMs can be initialized
by cloning the parameters of the trained single DBM. Nevertheless,
we observe that since the clustering layer is not needed during the
detection phase, we propose to remove the first visible layer $\visible^{\lrr 1}$
and the first hidden layer $\hiddenone$, converting a region DBM
to a RBM. This conversion helps $\modelDBM$ perform more efficiently
because updating the shallow networks of RBM with $\text{CD}{}_{1}$
is much faster than updating DBMs with Gibbs sampling and mean-field. 

Overall, the streaming version of $\modelDBM$ includes the following
steps of: i) using the single DBM parameters to initialize the region
DBMs; ii) keeping the biases and the connection matrix of reconstruction
layer $\hiddentwo$ and its corresponding visible layer $\visible^{\lrr 2}$
to form region RBMs; iii) reducing the number of hidden units to obtain
smaller RBMs using Alg. \ref{algRBMReduction}; iv) fine-tuning the
region RBMs using the corresonding patch data from the training videos;
and v) applying the same procedure in $\modelRBM$ to detect and update
the region RBMs. The steps i-iv) are performed in the training phase
as soon as the single DBM has been learned whilst the last step is
triggered in the detection phase. The step iii) is introduced because
the reconstruction layer in $\modelDBM$ usually needs more units
than the region RBMs in $\modelRBM$ with the same reconstruction
capacity. Therefore, we propose to decrease the number of DBM's hidden
units by discarding the units that have less contributions (low average
connection strength $\alpha_{n}$ in the line 7 of Alg. \ref{algRBMReduction})
to reconstruct the data before using the training set to fine-tune
these new RBMs.

\section{Experiment\label{sec:Experiment}}

In this section, we investigate the performance of our proposed $\modelEAD$,
wherein we demonstrate the capacity of capturing data regularity,
reconstructing scenes and detecting anomaly events. We provide a quantitative
comparison with state-of-the-art unsupervised anomaly detection systems.
In addition, we introduce some potential applications of our methods
for video analysis and scene clustering.

\begin{figure*}[th]
\begin{centering}
\includegraphics[width=0.8\textwidth]{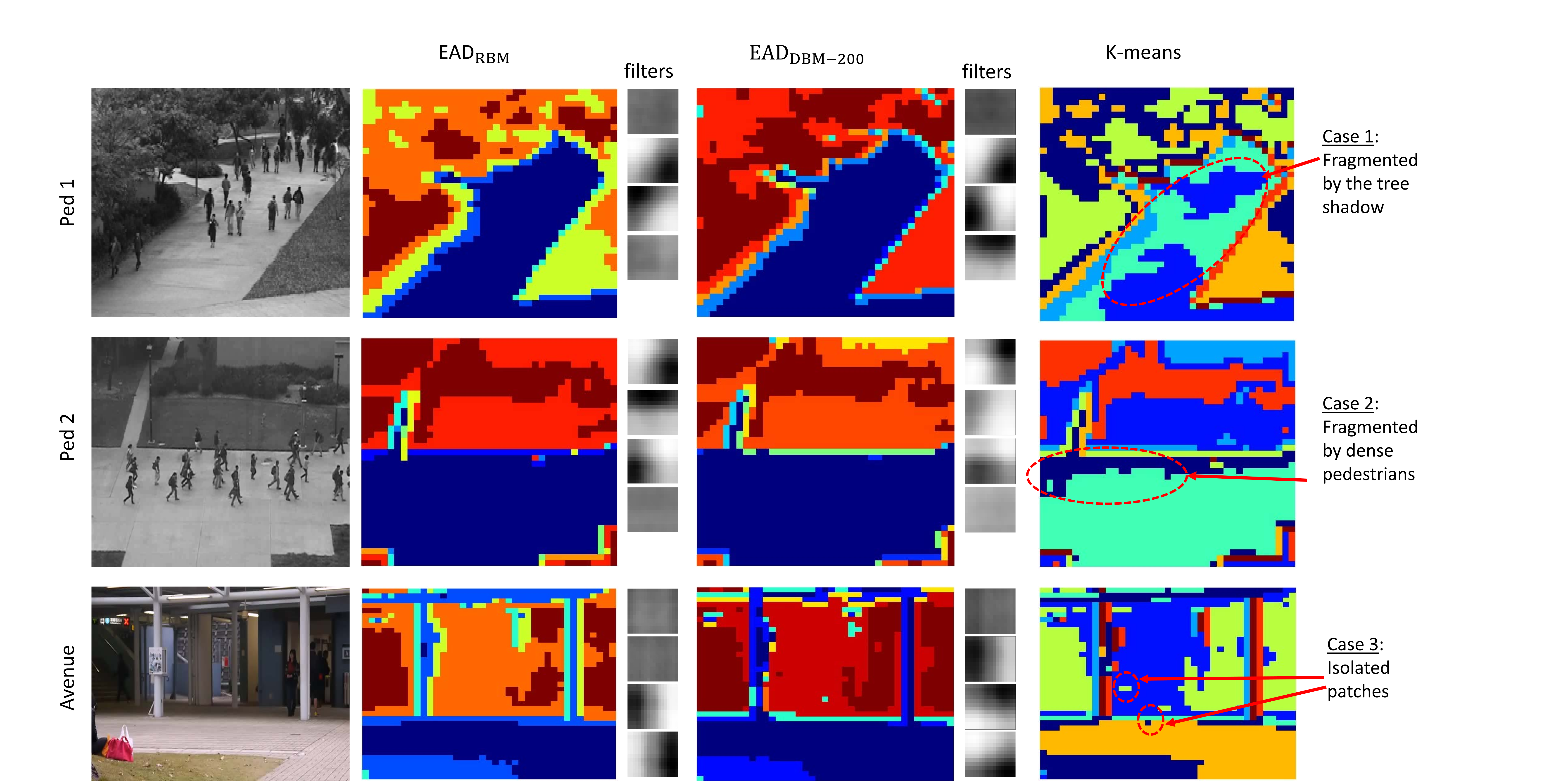}
\par\end{centering}

\caption{Clustering results of the clustering $\protect\modelRBM$, $\protect\modelDBMtwo$
and k-means in UCSD Ped 1, Ped 2 and Avenue datasets.}
\label{figclustering_results}
\end{figure*}

The experiments are conducted on 3 benchmark datasets: UCSD Ped 1,
Ped 2 \cite{Li.WeiXin_etal_2014PAMI_supervised_MDT} and Avenue \cite{Lu.Cewu_etal_2013ICCV_Supervised-Sparse}.
Since these videos are provided at different resolutions, we resize
all of them into the same frame size of $240\times360$. Following
the unsupervised learning setting, we discard all label information
in the training set before fitting the models. All methods are evaluated
on the testing videos using AUC (area under ROC curve) and EER (equal
error rate) at frame-level \cite{Li.WeiXin_etal_2014PAMI_supervised_MDT},
pixel-level \cite{Li.WeiXin_etal_2014PAMI_supervised_MDT} and dual-pixel
level \cite{Sabokrou.Mohammad_etal_2015CVPRW_DL_AE_GaussianClassifier}.
At frame-level, the systems only focus on answering whether a frame
contains any anomaly object or not. By contrast, pixel-level requires
the systems to take into account the locations of anomaly objects
in frames. A detection is considered to be correct if it covers at
least $40\%$ of anomaly pixels in the ground-truth. However, the
pixel-level evaluation can be easily fooled by assigning anomalous
labels to every pixels in the scene. Dual-pixel level tackles this
issue by adding one constraint of at least $\alpha$ percent of decision
being true anomaly pixels. It can be seen that pixel-level is a special
case of the dual-pixel level when $\alpha=0$.

\begin{figure}[t]
\centering{}\includegraphics[width=1\columnwidth]{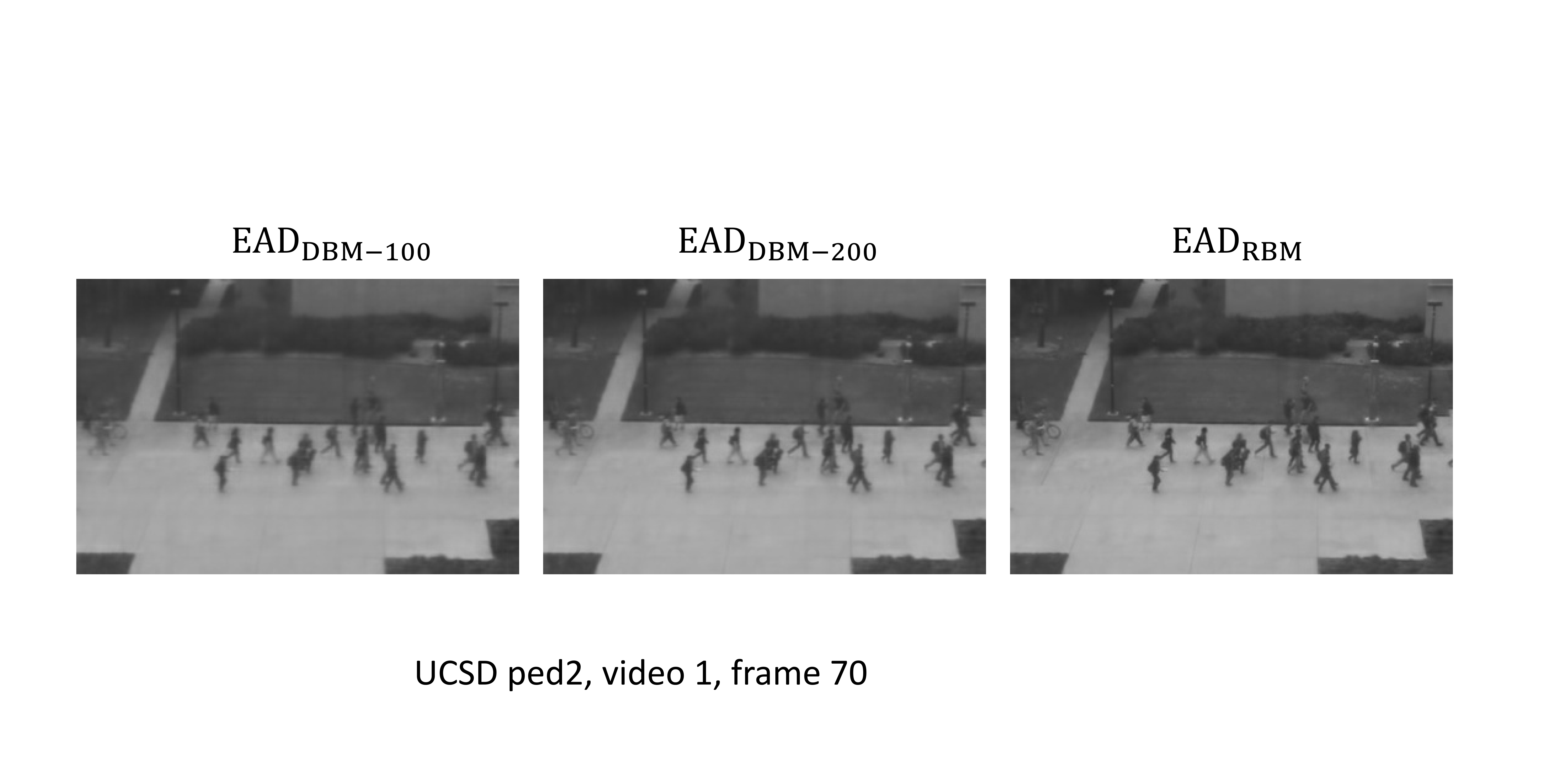}\caption{Reconstruction quality of $\protect\modelDBMone$, $\protect\modelDBMtwo$
and $\protect\modelRBM$. $\protect\modelDBMtwo$ has equivalent reconstruction
quality to $\protect\modelRBM$ whilst $\protect\modelDBMone$ produces
unclear image (e.g., the cyclist on the left of the scene). \label{figReconQuality}}
\end{figure}

To deal with the changes of objects in size and scale, we process
video frames at the scale ratios of $1.0$, $0.5$ and $0.25$ which
indicate no, a half and a quarter reduction in each image dimension.
We set the patch size to $12\times18$ pixels and patch strides to
$6$ and $9$ pixels in vertical and horizontal directions respectively.
For $\modelRBM$, we use a clustering RBM with 4 hidden units and
region RBMs with $100$ hidden units. All of them are trained using
$\text{CD}_{1}$ with $500$ epochs and a learning rate $\eta=0.1$.
For $\beta$ and $\gamma$, we tune these hyperparameters to achieve
the best balanced AUC and EER scores and come up with $\mbox{\ensuremath{\beta}\ = 0.0035 }$and
$\gamma=10$. For $\modelDBM$ system, a DBM with 4 hidden units in
the clustering layer and $200$ hidden units in reconstruction layer
(Fig. \ref{figTwoEndDBM}) is investigated. In fact, we also test
a DBM network with $\hiddenone$ of 4 units and $\hiddentwo$of $100$
units. However, since there exists correlations between these hidden
layers, $100$ hidden units in DBM cannot produce similar reconstruction
quality to $100$ hidden units in region RBMs (Fig. \ref{figReconQuality})
and therefore, more reconstruction units are needed in DBMs. As a
result, we use DBM with 200 reconstruction units in all our experiments.
We train DBMs using PCD \cite{Tieleman.Tijmen_2008ICML_PCD} with
$500$ epochs, pretraining procedure in \cite{Salakhutdinov.Ruslan_Hinton.Geoffrey_2009AISTATS}
with $50$ epochs and a smaller learning rate of $0.001$. Two thresholds
are $\beta=0.0043$ and $\gamma=10$. For the streaming versions of
$\modelRBM$ and $\modelDBM$ (we name them $\modelSRBM$ and $\modelSDBM$),
we split videos into non-overlapping chunks of $L=20$ contiguous
frames. After every frame, the systems update their parameters using
gradient ascent procedure in $20$ epochs. The thresholds $\beta$
and $\gamma$ are set to $0.003$ and $10$ respectively. All experiments
are conducted on a Linux server with 32 CPUs of 3 GHz and 126 GB RAM.

\begin{figure}[t]
\centering{}\includegraphics[width=1\columnwidth]{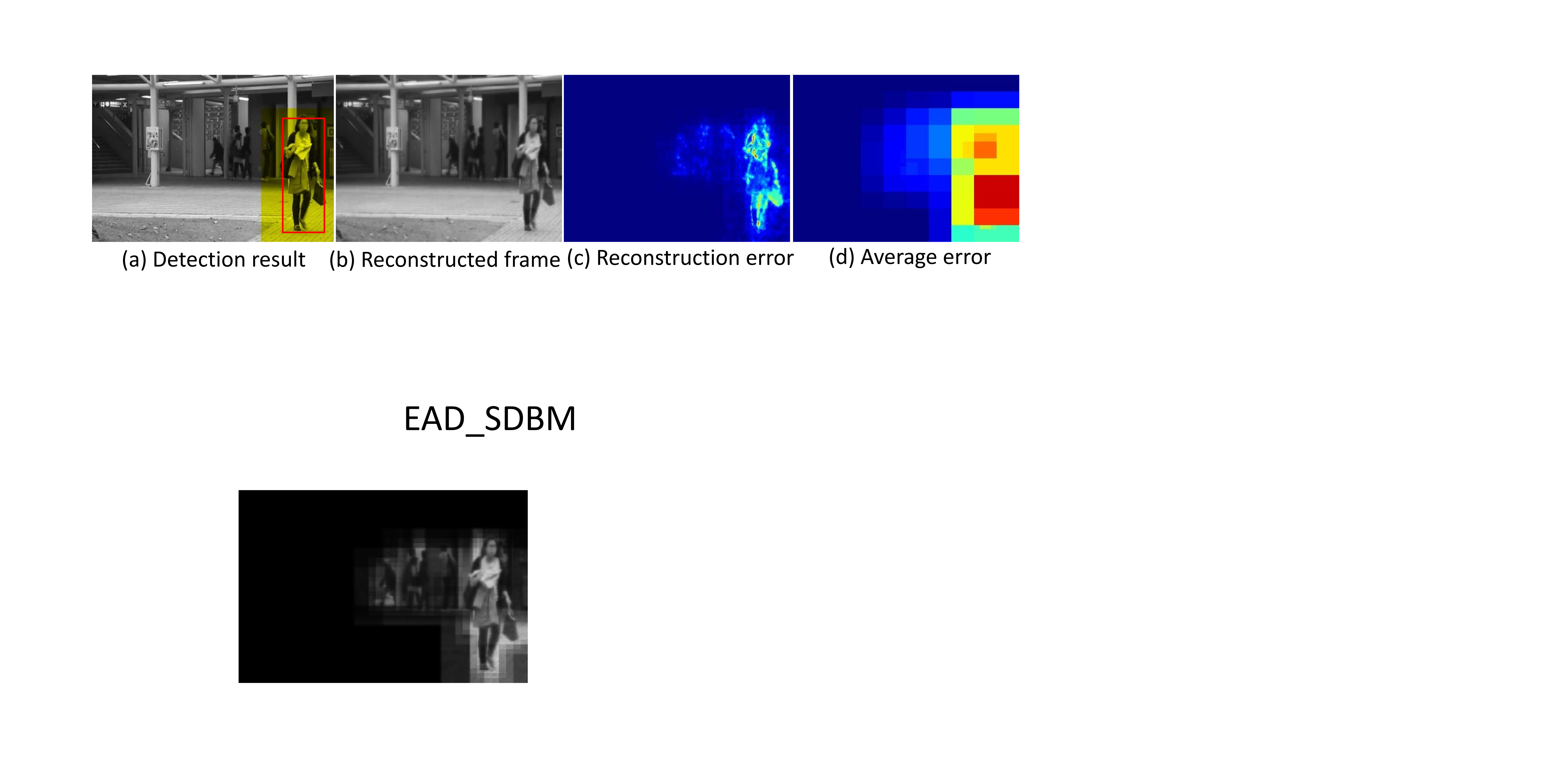}\caption{Data reconstruction produced by $\protect\modelSDBM$ on Avenue dataset:
(a) the original frame with a yellow region of detected outlier female
and a red rectangle of ground-truth, (b) reconstructed frame, (c)
reconstruction error image, (d) average reconstruction errors of patches.\label{figExample}}
\end{figure}

\begin{figure*}[th]
\centering{}\includegraphics[width=0.8\textwidth]{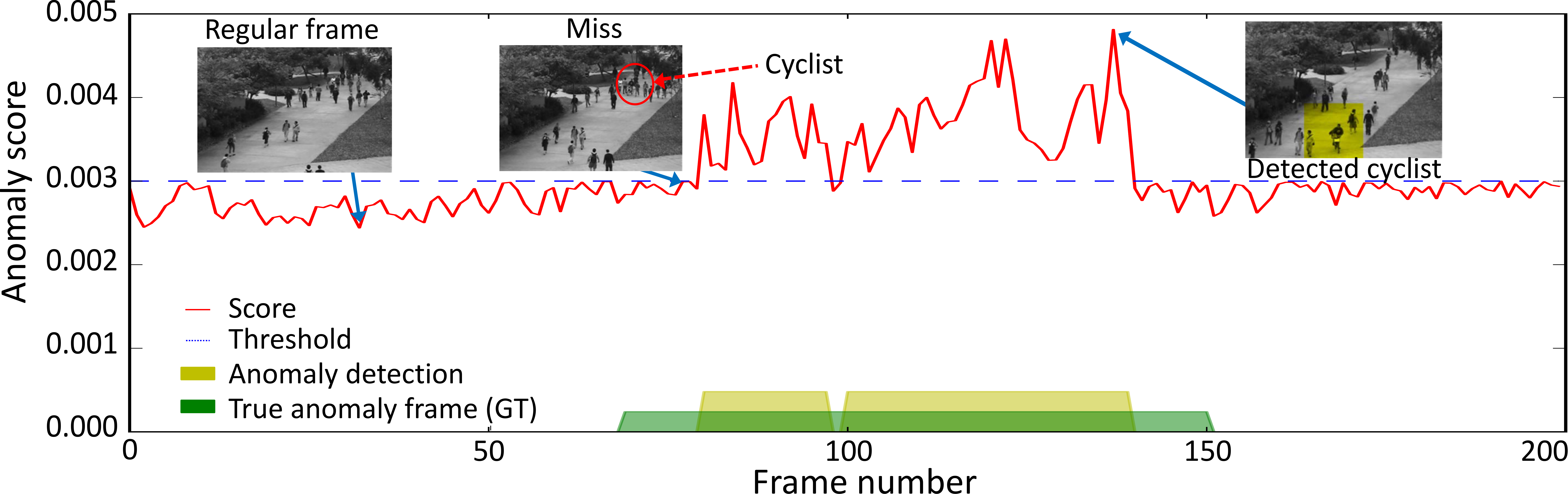}\vspace{-1em}
\caption{Average reconstruction error per frame in test video \#1 of UCSD Ped
1 dataset. The shaded green region illustrates anomalous frames in
the ground truth, while the yellow anomalous frames detected by $\protect\modelSDBM$.
The blue line shows the threshold.\label{figAnomScore}}
\vspace{-1em}
\end{figure*}

\subsection{Scene clustering\label{sub:Scene-clustering}}

In the first experiment, we investigate the performance of clustering
modules in the proposed systems. More specifically, a clustering RBM
network of $4$ hidden units has a responsibility to do the clustering
task in $\modelRBM$ while the first hidden layer of $4$ units is
used to do this step in $\modelDBM$. The clustering results are shown
in Fig. \ref{figclustering_results}. Overall, both $\modelRBM$ and
$\modelDBM$ discover plausible clustering maps of similar quality.
Using $4$ binary units, we expect that the systems can group video
scenes into maximum $2^{4}=16$ groups but interestingly, they use
less and return varied number of clusters depending on the video scenes
and scales. For examples, $\modelRBM$ uses ($6$, $7$, $10$) clusters
for three scales ($1.0$, $0.5$, $0.25$) respectively in Ped 1 dataset
whilst the numbers are ($9$, $9$, $8$) and ($6$, $9$, $9$) in
Ped 2 and Avenue datasets. Similarly, we observe the triples produced
by $\modelDBM$ are ($9$, $9$, $11$) in Ped 1, ($7$, $9$, $6$)
in Ped 2 and ($9$, $9$, $8$) in Avenue. The capacity of automatically
selecting the appropriate number of groups shows how well our $\modelEAD$s
can understand the scene and its structure.

 For further comparison, we deploy $k$-means with $k=8$ clusters,
the average number of clusters of $\modelRBM$ and $\modelDBM$ described
above. The clustering maps in the last column of Fig. \ref{figclustering_results}
show $k$-means fails to recognize large homogeneous regions, resulting
in fragmenting them into many smaller regions. This is due to the
impact of surrounding objects and complicated events in reality such
as the shadow of the trees (case 1 in the figure) or the dynamics
of crowded areas in the upper side of the footpath (case 2). In addition,
$k$-means tends to produce many spots with wrong labels inside large
clusters as shown in case 3. By contrast, two energy-based systems
consider the factor of uncertainty and therefore are more robust to
these randomly environmental factors. 

\begin{figure*}[t]
\centering{}\vspace{2mm}
\includegraphics[width=0.9\textwidth]{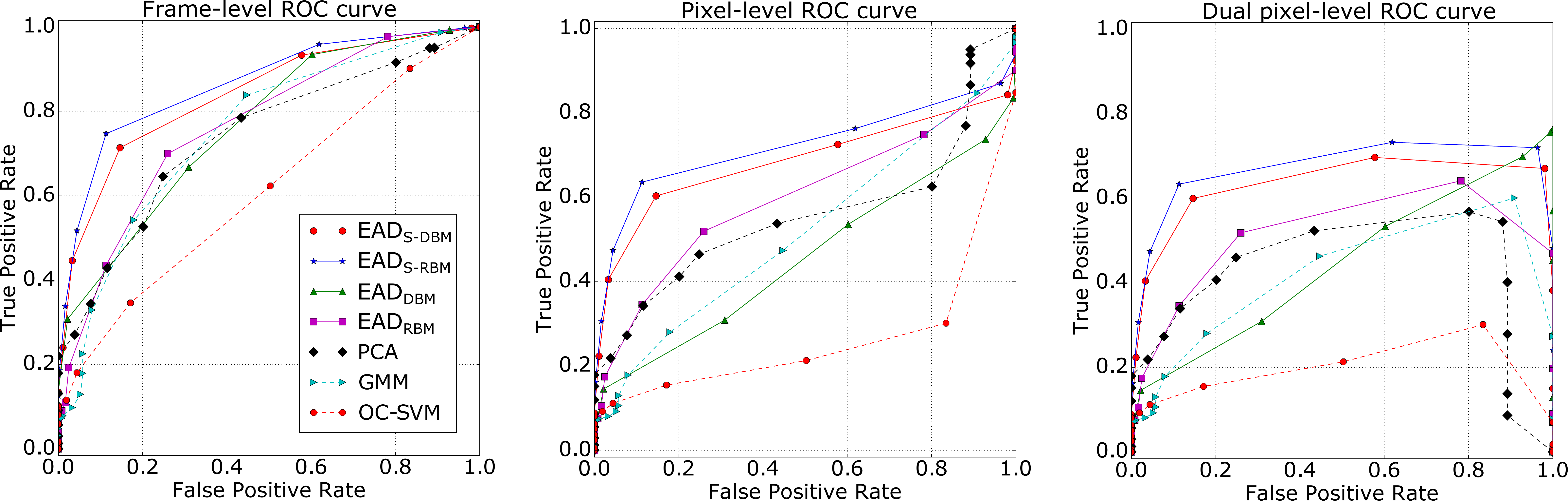}\vspace{-1mm}
\caption{Comparison ROC curves on UCSD Ped 2. Three figures share the same
legend. Higher curves indicate better performance. It is notable that,
unlike frame and pixel-level evaluations, dual-pixel level curves
may end at any points lower than (1,1). \label{figROC}}
\end{figure*}

\definecolor{grey}{rgb}{0.9,0.9,0.9}

\begin{table*}[th]
\begin{centering}
\vspace{1mm}

\par\end{centering}

\begin{centering}
\resizebox{1.0\textwidth}{!}{%%
\begin{tabular}{>{\raggedright}m{3.15cm}lr|rr|r|rr|rr|r|rr|rr|r}
\cline{2-16} 
\multirow{3}{3.15cm}{} & \multicolumn{5}{c|}{\textbf{Ped 1}} & \multicolumn{5}{c|}{\textbf{Ped 2}} & \multicolumn{5}{c}{\textbf{Avenue}}\tabularnewline
\cline{2-16} 
 & \multicolumn{2}{c|}{Frame} & \multicolumn{2}{c|}{Pixel} & Dual & \multicolumn{2}{c|}{Frame\textbf{ }} & \multicolumn{2}{c|}{Pixel} & Dual & \multicolumn{2}{c|}{Frame} & \multicolumn{2}{c|}{Pixel} & Dual\tabularnewline
 & AUC & EER & AUC & EER & AUC & AUC & EER & AUC & EER & AUC & AUC & EER & AUC & EER & AUC\tabularnewline
\hline 
\multicolumn{3}{l|}{\textbf{Unsupervised methods}} &  &  &  &  &  &  &  &  &  &  &  &  & \tabularnewline
PCA \cite{Pham.DucSon_etal_2011ICDM_principle_eigenvectors}  & 60.28 & 43.18 & 25.39 & 39.56 & 8.76 & 73.98 & 29.20 & 55.83 & 24.88 & 44.24 & 74.64 & 30.04 & \uline{52.90} & 37.73 & 43.74\tabularnewline
\rowcolor{grey}OC-SVM & 59.06 & 42.97 & 21.78 & 37.47 & 11.72 & 61.01 & 44.43 & 26.27 & 26.47 & 19.23 & 71.66 & 33.87 & 33.16 & 47.55 & 33.15\tabularnewline
GMM & 60.33 & 38.88  & 36.64  & 35.07 & 13.60 & 75.20 & 30.95 & 51.93 & 18.46 & 40.33 & 67.27 & 35.84 & 43.06 & 43.13 & 41.64\tabularnewline
\multicolumn{3}{l|}{\textbf{Deep models}} &  &  &  &  &  &  &  &  &  &  &  &  & \tabularnewline
CAE (FR) \cite{Ribeiro.Manasses_etal_2017PRL_DL_ConvAE} & 53.50 & 48.00 & \_ & \_ & \_ & 81.40 & 26.00 & \_ & \_ & \_ & 73.80 & 32.80 & \_ & \_ & \_\tabularnewline
\rowcolor{grey}ConvAE \cite{Hasan.Mahmudul_etla_2016CVPR_DL_ConvAE} & \textbf{81.00} & \textbf{27.90} & \_ & \_ & \_ & \textbf{90.00} & 21.70 & \_ & \_ & \_ & 70.20 & \textbf{25.10} & \_ & \_ & \_\tabularnewline
\hline 
\multicolumn{3}{l|}{\textbf{Our systems}} &  &  &  &  &  &  &  &  &  &  &  &  & \tabularnewline
$\modelRBM$ & 64.83 & 37.94 & 41.87 & 36.54 & 16.06 & 76.70 & 28.56 & 59.95 & 19.75 & 46.13 & 74.88 & 32.49 & 43.72 & 43.83 & 41.57\tabularnewline
\rowcolor{grey}$\modelDBM$(100 units)  & 64.33 & 39.42 & 26.96 & 34.93 & 19.24 & 71.63 & 34.38 & 38.82 & 20.50 & 37.65  & 77.40 & 30.96 & 43.86 & 45.21 & 43.15\tabularnewline
$\modelDBM$(200 units)  & 64.60 & 39.29 & 28.16 & 35.19 & \uline{20.21} & 76.52 & 32.04 & 45.56 & 19.40 & 44.17 & \uline{77.53} & 30.79 & 42.94 & 44.61 & 42.26\tabularnewline
\rowcolor{grey}$\modelSRBM$ & \uline{70.25} & \uline{35.40} & \textbf{48.87} & \textbf{33.31} & \textbf{22.07} & \uline{86.43} & \textbf{16.47} & \textbf{72.05} & \textbf{15.32} & \textbf{66.14} & \textbf{78.76} & \uline{27.21} & \textbf{56.08} & \textbf{34.40} & \textbf{53.40}\tabularnewline
$\modelSDBM$(200 units)  & 68.35 & 36.17 & \uline{43.17} & \uline{34.79} & 20.02 & 83.87 & \uline{19.25} & \uline{68.52} & \uline{17.16} & \uline{62.69} & 77.21 & 28.52 & 52.62 & \uline{36.84} & \uline{51.43}\tabularnewline
\end{tabular}}
\par\end{centering}

\begin{centering}

\par\end{centering}

\begin{centering}

\par\end{centering}

\begin{centering}
\vspace{2mm}

\par\end{centering}

\caption{Anomaly detection results (AUC and EER) at frame-level, pixel-level
and dual pixel-level ($\alpha=5\%)$ on 3 datasets. Higher AUC and
lower EER indicate better performance. Meanwhile, high dual-pixel
values point out more accurate localization. We do not report EER
for dual-pixel level because this number do not always exist. Best
scores are in bold whilst the next best is underlined. Note that the
frame-level results of CAE (FR) and ConvAE are taken from \cite{Ribeiro.Manasses_etal_2017PRL_DL_ConvAE}
and \cite{Hasan.Mahmudul_etla_2016CVPR_DL_ConvAE} respectively, but
the pixel-level and dual-pixel level results are not available.\label{tableResult}}
\vspace{-2em}
\end{table*}

\subsection{Scene reconstruction}

The key ingredient of our systems for distinguishing anomaly behaviors
in videos is the capacity of reconstructing data, which directly affects
detection results. In this part, we give a demonstration of the reconstruction
quality of our proposed systems. Fig. \ref{figExample} is an example
of a video frame with an anomaly object, which is a girl moving toward
the camera. Our $\modelSDBM$ produces the corresponding reconstructed
frame in Fig. \ref{figExample}b whilst the pixel error map and the
average error map are shown in Fig. \ref{figExample}c and \ref{figExample}d,
respectively. It can be seen that there are many high errors in anomaly
regions but low errors in the other regular areas. This confirms that
our model can capture the regularity very well and recognize unusual
events in frames using reconstruction errors (Fig. \ref{figExample}a). 

To demonstrate the change of the reconstruction errors with respect
to the abnormality in frame sequence, we draw the maximum average
reconstruction error in a frame as a function of frame index. As shown
in Fig. \ref{figAnomScore}, the video \#1 in UCSD Ped 1 starts with
a sequence of normal pedestrians walking on a footpath, followed by
an irregular cyclist moving towards the camera. Since the cyclist
is too small and covered by many surrounding pedestrians in the first
few frames of its emergence, its low anomaly score reveals that our
system cannot distinguish it from other normal objects. However, the
score increases rapidly and exceeds the threshold after several frames
and the system can detect it correctly. 

\begin{figure*}[th]
\begin{centering}
\includegraphics[width=0.8\textwidth]{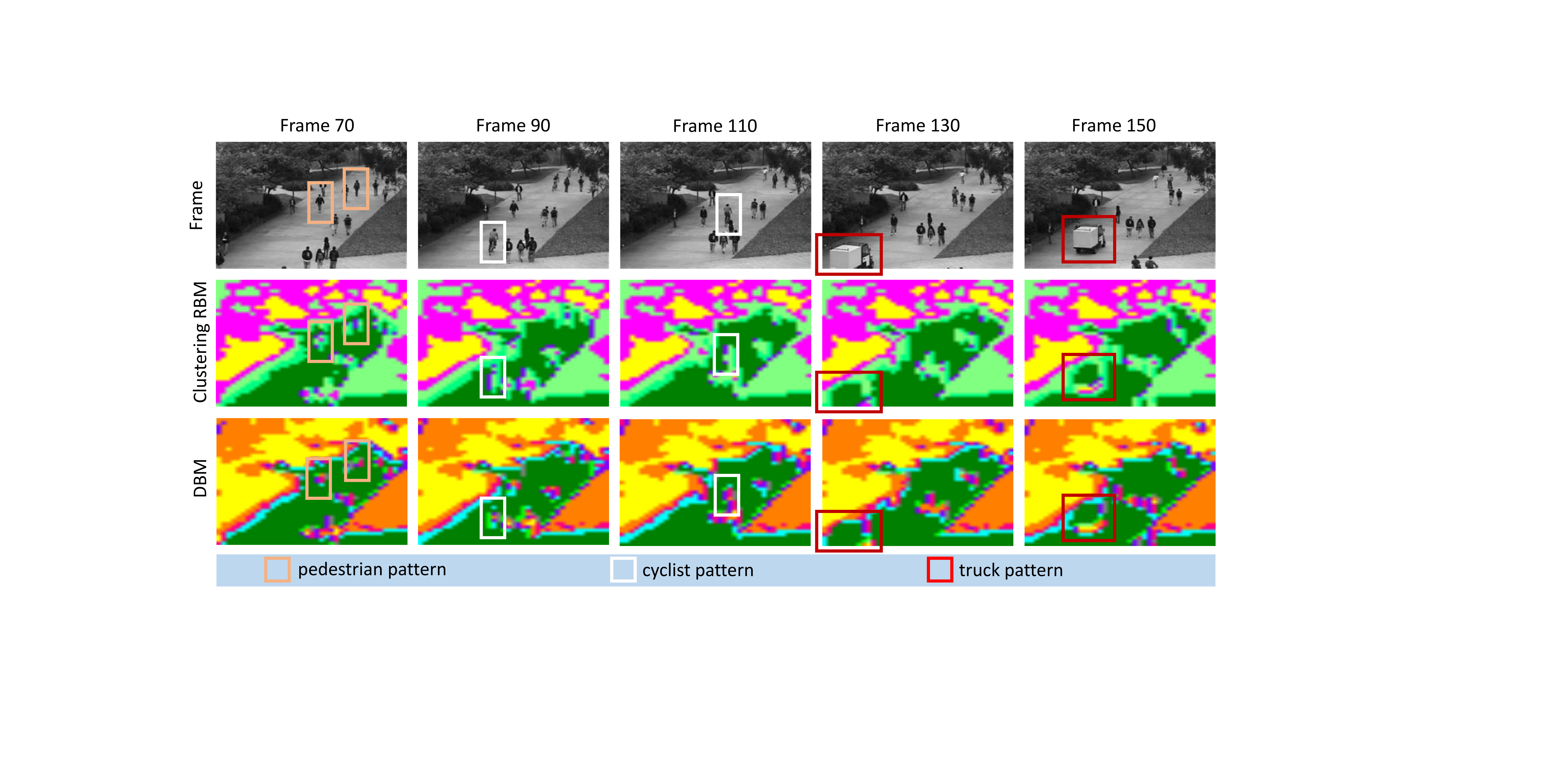}
\par\end{centering}

\caption{Some examples of pattern maps illustrate how $\protect\modelDBM$
expresses different objects at the abstract levels in a video sequence.
The frames were taken from video \#14, UCSD Ped 1 dataset.\label{figvideoanalysis}}

\vspace{-0.5cm}
\end{figure*}

\begin{figure*}[th]
\begin{centering}
\includegraphics[width=0.8\textwidth]{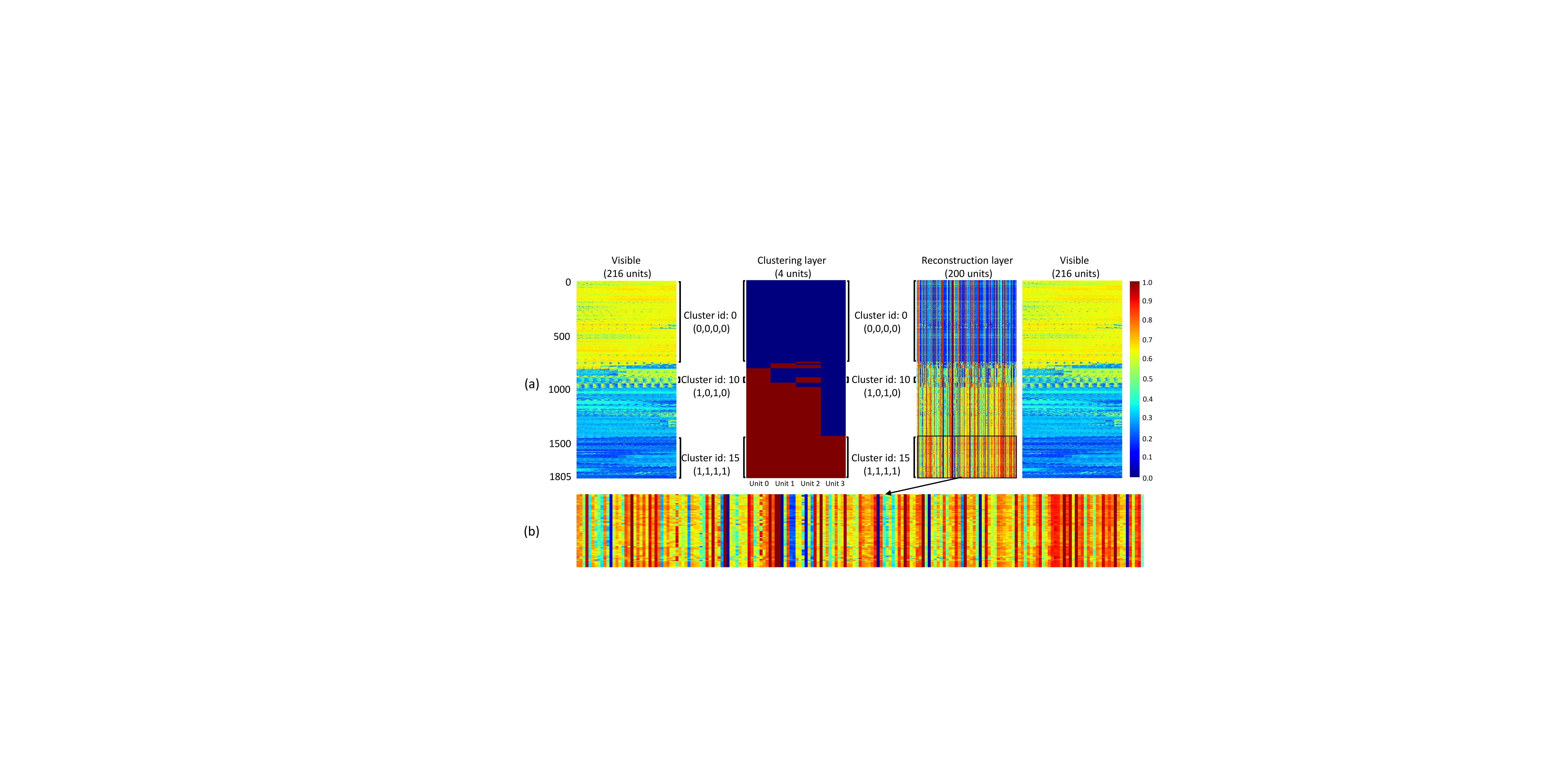}
\par\end{centering}

\caption{Activation visualization of the DBM network corresponding to 1805
patches of 5 random UCSD Ped 2 frames at scale 0.5: a) activations
of the layers and their correlations; and b) a close view of reconstruction
layer activations of different patches sharing the same cluster id.
\label{figlayeralignment}}

\vspace{-0.5cm}
\end{figure*}

\subsection{Anomaly detection}

To evaluate our $\modelEAD$ systems in anomaly detection task, we
compare $\modelRBM$ and $\modelDBM$ and our streaming versions $\modelSRBM$
and $\modelSDBM$ with several unsupervised anomaly detection systems
in the literature. These systems can be categorized into a) unsupervised
learning methods including Principal Component Analysis (PCA), One-Class
Support Vector Machine (OC-SVM) and Gaussian Mixture Model (GMM);
and b) the state-of-the-art deep models including CAE \cite{Ribeiro.Manasses_etal_2017PRL_DL_ConvAE}
and ConvAE \cite{Hasan.Mahmudul_etla_2016CVPR_DL_ConvAE}. 

We use the implementation of PCA with optical flow features for anomaly
detection in \cite{Pham.DucSon_etal_2011ICDM_principle_eigenvectors}.
For unsupervised baselines of OC-SVM and GMM, we use the same procedure
of our $\modelRBM$ framework but use $k$-means, instead of the clustering
RBM, to group image patches into clusters and OC-SVM/GMM models, instead
of the region RBMs, to compute the anomaly scores. Their hyperparameters
are turned to gain the best cross-validation results, namely we set
kernel width and lower bound of the fraction of support vectors to
$0.1$ and $10^{-4}$ for OC-SVM while the number of Gaussian components
and anomaly threshold in GMM are $20$ and $-50$ respectively. It
is worthy to note that we do not consider the incremental versions
of PCA, OC-SVM and GMM since it is not straightforward to update those
models in our streaming setting. Finally, the results of competing
deep learning methods are adopted from their original papers. Although
CAE and ConvAE were tested on both frame data and hand-crafted features
in \cite{Ribeiro.Manasses_etal_2017PRL_DL_ConvAE,Hasan.Mahmudul_etla_2016CVPR_DL_ConvAE},
we only include their experimental results on raw data for fair comparison
with our models which mainly work without hand-crafted features. 

\begin{table}[t]
\begin{centering}

\par\end{centering}

\begin{centering}
\resizebox{1.0\columnwidth}{!}{%%
\begin{tabular}{>{\raggedright}p{1.5cm}|rrr|r}
\multirow{1}{1.5cm}{} & Ped 1 & Ped 2 & Avenue & \textbf{Average}\tabularnewline
\hline 
\hline 
$\modelRBM$ & 137,736 & 79,576 & 122,695 & \textbf{113,336}\tabularnewline
$\modelDBM$ & 123,073 & 108,637 & 125,208 & \textbf{118,973}\tabularnewline
\end{tabular}}
\par\end{centering}

\vspace{0.2cm}

\caption{Training time of $\protect\modelRBM$ and $\protect\modelDBM$ in
second.\label{tableTrainingtime}}
\end{table}

Table \ref{tableResult} reports the experimental results of our systems
versus all methods whilst Fig. \ref{figROC} shows ROC curves of our
methods and unsupervised learning methods. Overall, our energy-based
models are superior to PCA, OC-SVM and GMM in terms of higher AUC
and lower ERR. Interestingly, our higher AUCs in dual-pixel level
reveals that our methods can localize anomalous regions more correctly.
These results are also comparable with other state-of-the-art video
anomaly detection systems using deep learning techniques (i.e., CAE
\cite{Ribeiro.Manasses_etal_2017PRL_DL_ConvAE} and ConvAE \cite{Hasan.Mahmudul_etla_2016CVPR_DL_ConvAE}).
Both CAE and ConvAE are deep Autoencoder networks (12 layers) that
are reinforced with the power of convolutional and pooling layers.
By contrast, our systems only have a few layers and basic connections
between them but obtain respectable performance. For this reason,
we believe that our proposed framework of energy-based models is a
promising direction to develop anomaly detection systems in future
surveillance applications. 

Comparing between $\modelRBM$ and $\modelDBM$, Table \ref{tableResult}
shows that $\modelDBM$ with 100 reconstruction hidden units is not
so good as $\modelRBM$ (with the same number of hidden units). This
is because the reconstruction units in DBMs have to make additional
alignment with the clustering units and therefore there is a reduction
in reconstruction and detection quality. However, by adding more units
to compensate for such decrease, our $\modelDBM$ with 200 hidden
units can obtain similar detection results to $\modelRBM$. Therefore,
we choose the DBM network with 200 reconstruction hidden units as
the core of our $\modelDBM$ system. To shorten notation, we write
$\modelDBM$ (without the explicit description of the number of hidden
units) for a system with 200 reconstruction hidden units.

The training time of two systems is reported in Table \ref{tableTrainingtime}.
Overall, there is no much different in training time between them
because DBM learning procedure with expensive Gibbs sampling and mean-field
steps and additional pretraining cost is more time-consuming than
$\text{CD}_{1}$ in RBM training. However, one advantage of $\modelDBM$
system is that it requires to train one DBM model for every video
scale versus many models (i.e., 9 models in average) in $\modelRBM$.
Another benefit of $\modelDBM$ is the capacity of model explanation,
which will be discussed in the following section.

\subsection{Video analysis and model explanation}

The clustering module in our systems is not only applied for scene
segmentation but also useful for many applications such as video analysis
and model explanation. Unlike other clustering algorithms that are
mainly based on the common characteristics (e.g., distance, density
or distribution) to group data points together, the clustering modules
in $\modelEAD$ leverage the representation power of energy-based
models (i.e. RBMs and DBMs) at abstract levels. For example, we understand
that a RBM with sufficient hidden units is able to reconstruct the
input data at excellent quality \cite{LeRoux.Nicolas_Bengio.Yoshua_2008NeuralComp_theory_addhidden}.
If we restrict it to a few hidden neurons, e.g., 4 units in our clustering
RBM, the network is forced to learn how to describe the data using
limited capacity, e.g., maximum 16 states for 4 hidden units, rendering
the low-bit representation of the data. This low-bit representation
offers an abstract and compact view of the data and therefore brings
us high-level information. Fig. \ref{figvideoanalysis} reveals abstract
views (pattern maps) of several frames from USCD Ped 1 dataset, which
are exactly produced by the clustering RBM in $\modelRBM$ and the
clustering layer in $\modelDBM$. It can be seen that different objects
have different patterns. More specifically, all people can be represented
as patterns of purple and lime blocks (frame 70 in Fig. \ref{figvideoanalysis})
but their combination varies in human pose and size. The variation
in the representation of people is a quintessence of articulated objects
with the high levels of deformation. On the other hand, a rigid object
usually has a consistent pattern, e.g., the light truck in frames
130 and 150 of Fig.~\ref{figvideoanalysis} has a green block to
describe a cargo space and smaller purple, yellow and orange blocks
to represent the lower part. This demonstration shows a potential
of our systems for video analysis, where the systems assist human
operators by filtering out redundant information at the pixel levels
and summarizing the main changes in videos at the abstract levels.
The pattern maps in Fig. \ref{figvideoanalysis} can also be used
as high level features for other computer vision systems such as object
tracking, object recognition and motion detection. 

The abstract representation of the videos also introduces another
nice property of model explanation in our systems. Unlike most video
anomaly detection systems \cite{Hasan.Mahmudul_etla_2016CVPR_DL_ConvAE,Ribeiro.Manasses_etal_2017PRL_DL_ConvAE,Xu.Dan_etal_2017CVIU_DL_StackedDenoisingAE_OCSVM,Medel.JeffersonRyan_Savakis.AndreasE_2016CoRR_DL_LSTM,Chong.YongShean_Tay.YongHaur_2017AdvNN_DL_ConvAE_ConvLSTM,Luo.W_etal_2017ICME_DL_ConvLSTMAE,Tran.HTM_Hogg.DC_2017BMVC_DL_ConvWinnerTakeAllAE,Ravanbakhsh.Mahdyar_etal_2017ICIP_DL_GAN,Sabokrou.Mohammad_etal_2016CVIU_CNN}
that only produce final detection results without providing any evidence
of model inference, the pattern maps show how our models view the
videos and therefore they are useful cues to help developers debug
the models and understand how the systems work. An example is the
mis-recognitions of distant cyclists to be normal objects. By examining
the pattern maps of frames 90 and 110 in Fig.~\ref{figvideoanalysis},
we can easily discover that distant cyclists share the same pattern
of purple and lime colors with pedestrians. Essentially, cyclists
are people riding bicycles. When the bicycles are too small, they
are unable to be recognized by the detectors and the cyclists are
considered as pedestrians. This indicates that our pattern maps can
offer a rational explanation of the system mistakes. 

There unlikely exists a model explanation capacity mentioned above
in $\modelRBM$ because its clustering module and its reconstruction
module are built separately and thus it does not ensure to obtain
an alignment between abstract representation (provided by clustering
RBMs) and detection decision (by region RBMs). As a result, what we
see in the pattern maps may not reflect what the model actually does.
By contrast, both clustering layer and reconstruction layer are trained
parallelly in $\modelDBM$, rendering a strong correlation between
them via their weight matrix. Fig. \ref{figlayeralignment} demonstrates
this correlation. We firstly collect all 1805 patches at the scale
0.5 from 5 random frames of UCSD Ped 2 dataset and then feed them
into the network and visualize the activation values of the layers
after running the mean-field procedure. Each picture can be viewed
as a matrix of (\# patches) rows and (\# units) columns. Each horizontal
line is the response of neurons and layers to the corresponding input
patch. As shown in Fig. \ref{figlayeralignment}a, there is a strong
agreement in color between the layers, for example, the cyan lines
in two visible layers always correspond to red lines in the clustering
layer and yellow lines in the reconstruction layer and similarly yellow
inputs are frequently related to the blue responses of the hidden
neurons. We can understand this by taking a closer look at the structure
of our proposed DBM. The connections with data ensure that the clustering
layer and the reconstruction layer have to represent the data whilst
their connections force them to align with each other.  However,
it is worthy to note that the reconstruction layer is not simply a
copy of the clustering layer but it adds more details towards describing
the corresponding data. As a result, there are still distinctions
between reconstruction layer responses of two different patches with
the same clustering layer responses.  Imagine that we have two white
patches of a footpath with and without some parts of a pedestrian.
As we know in Sec. \ref{sub:Scene-clustering}, these patches are
assigned to the same cluster or have the same clustering layer states
that represent footpath regions. Next, these states specify the states
of the reconstruction layer and make them similar. However, since
these patches are different, the patch with the pedestrian slightly
modifies the state of the reconstruction layer to describe the presence
of the pedestrian. Fig. \ref{figlayeralignment}b confirms this idea.
All reconstruction layer responses have the same cluster layer state
of $\lrr{0,0,0,0}$, and therefore the similar horizontal color strips,
but they are still different in intensity. All aforementioned discussions
conclude that the clustering layer in DBM is totally reliable to reflect
the operation of the system and it is useful to visualize and debug
the models. It is noteworthy that this capacity is not present in
shallow networks like RBMs.

\section{Conclusion\label{sec:Conclusion}}

This study presents a novel framework to deal with three existing
problems in video anomaly detection, that are the lack of labeled
training data, no explicit definition of anomaly objects and the dependence
on hand-crafted features. Our solution is based on energy-based models,
namely Restricted Boltzmann Machines and Deep Boltzmann Machines,
that are able to learn the distribution of unlabeled raw data and
then easily isolate anomaly behaviors in videos. We design our anomaly
detectors as 2-module systems of a clustering RBM/layer to segment
video scenes and region RBMs/reconstruction layer to represent normal
image patches. Anomaly signals are computed using the reconstruction
errors produced by the reconstruction module. The extensive experiments
conducted in 3 benchmark datasets of UCSD Ped 1, Ped 2 and Avenue
show the our proposed framework outperforms other unsupervised learning
methods in this task and achieves comparable detection performance
with the state-of-the-art deep detectors. Furthermore, our framework
also has a lot of advantages over many existing systems, i.e. the
nice capacities of scene segmentation, scene reconstruction, streaming
detection, video analysis and model explanation. 

\bibliographystyle{IEEEtran}
\bibliography{papers_short}

\vspace{-1cm}

\begin{IEEEbiography}[{\includegraphics[width=1in]{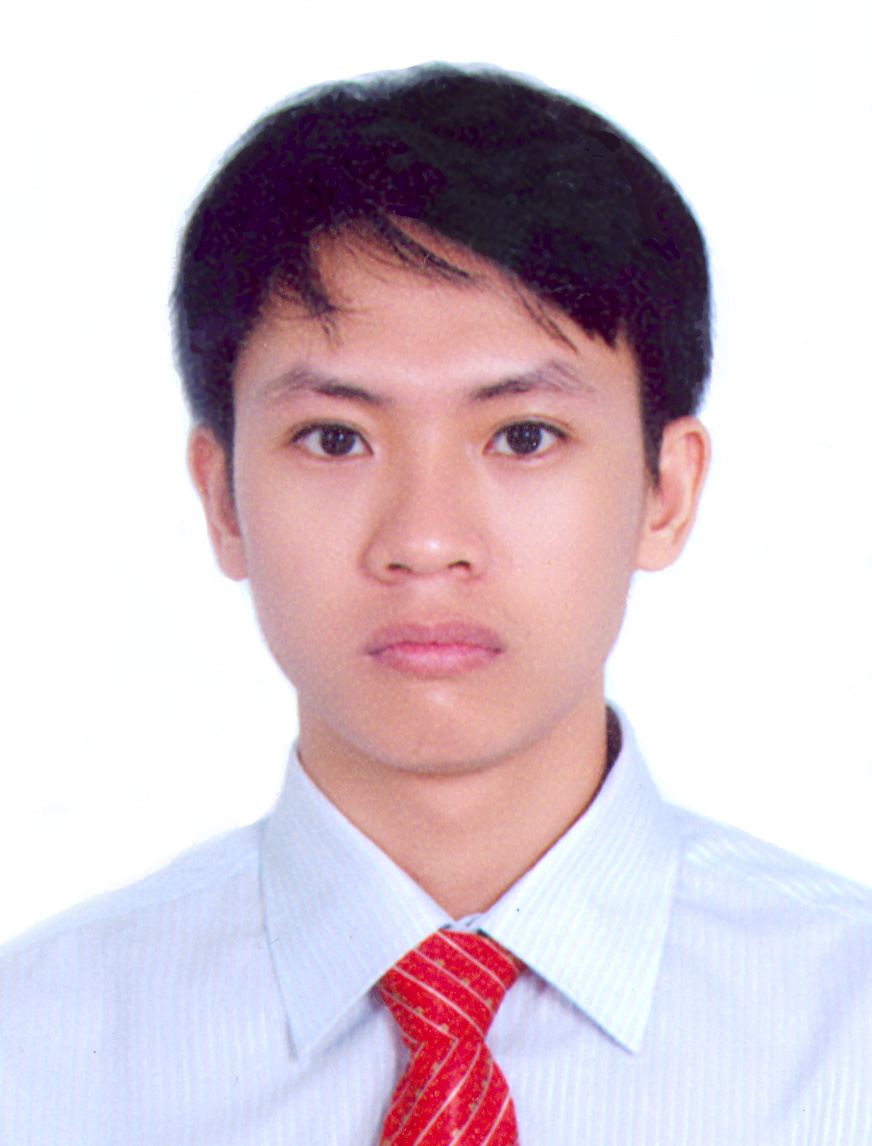}}]{Hung
Vu} is a PhD candidate and Postgraduate Research Scholar supported
by Deakin University, Australia. He received his Bachelor and Master
of Computer Science degrees from University of Sciences, HCMC, Vietnam
National University (VNU), Vietnam, in 2008 and 2011. His research
interests are the intersection of computer vision and machine learning.
His current projects focus on deep generative networks, Boltzmann
machines and their applications to anomaly detection and video surveillance. 

\end{IEEEbiography}

\vspace{-1cm}

\begin{IEEEbiography}[{\includegraphics[width=1in]{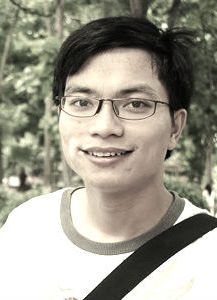}}]{Tu
Dinh Nguyen} obtained a Bachelor of Science from Vietnam National
University, and a PhD in Computer Science from Deakin University in
2010 and 2015, respectively. Currently he is a Research Fellow at
Deakin University, Australia. His research interests are deep generative
models, kernel methods, online learning and distributed computing.
He has won several research awards: Best Application Paper Award at
PAKDD conference (2017), Best Runner-Up Student Paper Award at PAKDD
conference (2015), and Honorable Mention Application Paper Award at
DSAA conference (2017). Tu achieved the title of ``Kaggle master''
on July 2014.\end{IEEEbiography}

\vspace{-1cm}

\begin{IEEEbiography}[{\includegraphics[width=1in]{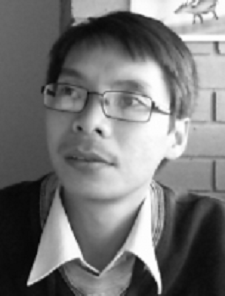}}]{Dinh
Phung} is currently a Professor and Deputy Director in the Centre
for Pattern Recognition and Data Analytics within the School of Information
Technology, Deakin University, Australia. He obtained a Ph.D. from
Curtin University in 2005 in the area of machine learning and multimedia
computing. His current research interest include machine learning,
graphical models, Bayesian nonparametrics, statistical deep networks,
online learning and their applications in diverse areas such as pervasive
healthcare, autism and health analytics, computer vision, multimedia
and social computing. He has published 180+ papers and 2 patents in
the area of his research interests. He has won numerous research awards,
including the Best Paper Award at PAKDD conference (2015), Best Paper
Award Runner-Up at UAI conference (2009), Curtin Innovation Award
from Curtin University for developing groundbreaking technology in
early intervention for autism (2011), Victorian Education Award (Research
Engagement 2013) and an International Research Fellowship from SRI
International in 2005. His research program has regularly attracted
competitive research funding from the Australian Research Council
(ARC) in the last 10 years and from the industry. He has delivered
several invited and keynote talks, served on 40+ organizing committees
and technical program committees for topnotch conferences in machine
learning and data analytics, including his recent role as the Program
Co-Chair for the Asian Conference on Machine Learning in 2014.\end{IEEEbiography}

\end{document}